\documentclass[review]{elsarticle}

\usepackage{lipsum} 

\usepackage{etoolbox}

\usepackage{lineno,hyperref}
\usepackage{bbm}
\modulolinenumbers[5]

\journal{Machine Learning Applications}

\usepackage{dirtree}
\usepackage{xcolor}
\usepackage{multirow}
\usepackage{subfigure}

\biboptions{authoryear}

\patchcmd{\pprintMaketitle}
 {\hrule}
 {\clearpage\hrule}
 {}{}
\appto\endfrontmatter{\clearpage}

\begin{document}

\begin{frontmatter}

\title{A Study of the Generalizability of Self-Supervised Representations}



\author[mymainaddress]{Atharva Tendle}
\ead{atharva.tendle@huskers.unl.edu}

\author[mymainaddress]{Mohammad Rashedul Hasan\corref{mycorrespondingauthor}}

\cortext[mycorrespondingauthor]{Corresponding author}
\ead{hasan@unl.edu}

\address[mymainaddress]{Department of Computer Science and Engineering, University of Nebraska-Lincoln, NE, USA}

\pagebreak

\begin{abstract}

Recent advancements in self-supervised learning (SSL) made it possible to learn generalizable visual representations from unlabeled data. The performance of Deep Learning models fine-tuned on pretrained SSL representations is on par with models fine-tuned on the state-of-the-art supervised learning (SL) representations. Irrespective of the progress made in SSL, its generalizability has not been studied extensively. In this article, we perform a deeper analysis of the generalizability of pretrained SSL and SL representations by conducting a domain-based study for transfer learning classification tasks. The representations are learned from the ImageNet source data, which are then fine-tuned using two types of target datasets: similar to the source dataset, and significantly different from the source dataset. We study generalizability of the SSL and SL-based models via their prediction accuracy as well as prediction confidence. In addition to this, we analyze the attribution of the final convolutional layer of these models to understand how they reason about the semantic identity of the data. We show that the SSL representations are more generalizable as compared to the SL representations. We explain the generalizability of the SSL representations by investigating its invariance property, which is shown to be better than that observed in the SL representations.

\end{abstract}


\end{frontmatter}

\section{Introduction}
\label{chap:intro}

Learning expressive visual representations from raw pixels is a challenging task. Deep Learning (DL) techniques make this possible by employing a hierarchical information processing system \citep{Goodfellow:2016}. The convolutional neural network (CNN) based DL models create increasingly informative representations of the input data using its layered architecture. The dominant DL paradigm to learn expressive representations is supervised learning (SL), which has been very successful in solving various vision tasks \citep{dong:2021} including image classification \citep{krizhevsky:2012, simonyan:2015, he:2016, tan:2019}, object detection \citep{girshick:2014, redmon:2016, he:2017}, segmentation \citep{minaee:2021, long:2015, he:2017, chen:2018, zoph:2020}, tracking \citep{wangz:2020, zhang:2020}, background subtraction \citep{bouwmans:2019}, and generative models \citep{kingma:2014, goodfellow:2014, parmar:2018, karras:2019}. However, this approach requires a large amount of task-specific labeled data. A data-efficient solution to this problem is transfer learning \citep{yosinski:2014}. In transfer learning, knowledge gained from the source domain is applied to a different but related target domain \citep{pan:2009}. This learning approach is efficient as it shifts the target data-based training of the DL models to the source data-based pretraining phase. More specifically, transfer learning consists of two phases: pretraining and fine-tuning. During the pretraining phase, a DL model is trained once on a large and generic source dataset to learn \textbf{generalizable} representations. Then, during the fine-tuning phase, these representations are adapted to the target domain based on the target data. The benefit of the transfer learning approach is that if the pretraining representations are generic enough, then the pretrained model can be fine-tuned using small target data for solving downstream tasks such as classification \citep{sharif:2014, oquab:2014, kolesnikov-bit:2019}. 





The standard approach for creating general-purpose pretrained representations is SL, which uses a large and generic labeled dataset such as ImageNet \citep{Russakovsky:2015} to pretrain a DL model. The data labels are used to compute prediction error that the pretraining model reduces by minimizing a loss function via the gradient descent algorithm. Although pretraining representations using the SL approach has been the de facto standard for transfer learning in computer vision applications \citep{ren:2016, chen:2017, weinzaepfel:2013, carreira:2016}, this approach suffers from two key limitations. It is \textit{inefficient}, and it limits the \textit{generalizability} of the representations.

The SL-based transfer learning approach is \textbf{inefficient} because it requires a large amount of labeled data \citep{zeiler:2014, huh:2016, mahajan:2018, cui:2018, kornblith:2019, he:2019}. There are various ways to attain these annotations such as from class labels \citep{Russakovsky:2015}, hashtags \citep{mahajan:2018}, bounding boxes \citep{everingham:2015, lin:2014}, etc. But manually annotating a large source dataset is expensive. The SL-based approach limits the \textit{generalizability} of transfer learning in at least two ways. First, the semantic annotations that are predefined often scale poorly to the long tail of visual concepts \citep{van:2017}. Second, there is an inherent bias in the learned representations \citep{dosovitskiy:2015}. The bias is caused by the technique that the model uses to reduce the loss between the predictions and the labels of the source data. More specifically, it does so by minimizing the SL cross-entropy loss function \citep{Goodfellow:2016}, which attracts samples from the same class and repels samples from other classes. This leads to representations that are mainly class or domain dependent. Hence, they might transfer poorly to other domains \citep{cui:2018, kornblith:2019, he:2019}. Generalizability suffers more when the new domain is different from the source domain \citep{dosovitskiy:2015, misra:2016, stock:2018}.

The root cause of the inefficiency and limited generalizability of the SL approach is attributed to its data-label based learning. Recent advancements in unsupervised learning have emerged as a promising solution to overcome both issues. More specifically, a sub-field of unsupervised learning known as the \textbf{self-supervised learning (SSL)} has become competitive to the state-of-the-art (SOTA) performance achieved by the SL techniques \citep{chen:2020v1, chen:2020v2, caron:2021, grill:2020, goyal:2021, zbontar:2021, chen:2021}. SSL obtains feedback signals from the underlying structure of data instead of explicit labels. In general, a SSL technique learns representations by predicting a hidden property of the input from the observable properties \citep{LeCun:2021}. It employs various strategies to generate pseudo labels in a semi-automatic fashion, which is motivated by the SSL approach in Natural Language Processing (NLP) \citep{Collobert:2008, mikolov:2013, pennington:2014, devlin:2018}.


Irrespective of the progress made by SSL-based pretrained representations for transfer learning in image recognition tasks \citep{chen:2020v1, chen:2020v2, caron:2021, goyal:2021, chen:2021}, no effort has been made to understand the generalizability of the SSL representations in a domain-specific way. Good generalization performance requires learning \textbf{invariant representations} such that the model's recognition performance remains unchanged in presence of variability in the high-dimensional images \citep{Sohn:2012}. Although the SSL techniques, which exhibit transfer learning performance on par with the SL based approach, are able to learn invariant representations \citep{caron:2021, chen:2020v2, he:2020}, it is not clear how the invariance property of the SSL representations differ from that of the SL representations. So far, the invariance property of the SSL representations is evaluated indirectly via their generalizability, which was determined by their prediction accuracy in transfer learning classification tasks \citep{caron:2021, chen:2020v2, he:2020}. We argue that prediction accuracy is not a suitable measure to understand the level of invariance in a model's representations. Two models with similar prediction accuracy on unseen data may have varied confidence in their predictions. Better invariance should translate into increased convictions in predictions when exposed to highly variable input. Thus, to develop a deep understanding of the invariance property of SSL representations, it is important to consider \textbf{prediction confidence} for target domains that are significantly different from the source domain. 

Once we understand the nature of invariance present in SSL representations in comparison to the SL representations, we need to \textbf{explain} the variability in their invariance property. However, no interpretability study has been done to understand the reasoning process of SL-based models. For example, we need to investigate when an SSL-based model makes a decision about the semantic identity of an image, what part of the input image it uses for decision making, and whether an SSL-based model's focusing ability is related to the invariance of its representations. Finally, unlike the dominant SOTA SL pretraining approach for representation learning \citep{kolesnikov-bit:2019}, there exist diverse SSL techniques for creating generalizable representations \citep{chen:2020v1, chen:2020v2, caron:2021, zbontar:2021, xiong:2021}. However, no study has been done to determine how the main SSL approaches vary in effectiveness (generalizability performance) and efficiency (fine-tuning time) when the target domain is noticeably diverse from the source domain used to pretrain the representations using various SSL techniques.



In this article, we conduct an in-depth analysis of the generalizability and invariance property of the visual representations created by various SSL techniques for transfer learning classification tasks. To determine the domain-specific generalizability of the SSL techniques, we systematically apply SSL techniques (i.e., SSL-based representation to fine-tune classifiers) on two different types of target datasets. More specifically, we use pretrained representations created from a single source dataset (i.e., ImageNet \citep{Russakovsky:2015}) to perform classification on the following two types of target datasets: similar to the source dataset, and significantly different from the source dataset. For the first type, we use the CIFAR datasets \citep{Krizhevsky:2010} due to their similarity in the categories and data distribution with the source dataset. For the second target domain, we use the camera-trap dataset. Camera-traps are motion-activated cameras used widely by ecologists to collect various information on wildlife populations such as habitat \citep{driscoll:2017}, population dynamics \citep{connell:2010}, and prey vigilance \citep{cherry:2015}. The camera-trap images exhibit large variability in the scale and location of the animals, the background and global illumination, and the distribution of the classes \citep{norouzzadeh:2018}. Concretely, we study (explore) the generalizability of the SSL representations, and try to understand (explain) the generalizability via their invariance property and the reasoning process. We investigate the following research questions.

\begin{itemize}
    \item RQ1 (exploratory): Do SSL representations exhibit better generalizability than the SL representations when the target domain is significantly different from the source domain?

    \item RQ2 (exploratory): Do the effectiveness (accuracy) and efficiency (epochs to reach convergence) of the main SSL approaches vary in a domain-specific way?

    \item RQ3  (explanatory): Do the SSL-based models exhibit better invariance property with respect to various transformations such as translation, rotation, flip, scale change as well as variation in background and global illumination?

    \item RQ4 (explanatory): How do the SSL-based models reason about the semantic identity of the data as compared to the SL-based model?

\end{itemize}


%
%

To address these questions, we design a set of studies that uses diverse target datasets as well as various SSL techniques. The SSL representations are pretrained using the ImageNet dataset \citep{Russakovsky:2015}, following which the downstream classification task is performed on datasets that share ImageNet like classes and class distributions (e.g., CIFAR-10 and CIFAR-100 \citep{Krizhevsky:2010}) as well as on datasets that are significantly different from ImageNet, i.e., camera-trap datasets (e.g., Snapshot Serengeti \citep{Swanson:2015}). We experiment with two main SSL approaches to capture the variance across different techniques: instance-based (i.e., SimCLR \citep{chen:2020v1, chen:2020v2} and Barlow Twins \citep{zbontar:2021}) and clustering-based (i.e., DeepClusterv2 \citep{caron:2021} and SwAV \citep{caron:2021}).

Our Main Contributions are as follows:

\begin{itemize}

  \item We provide a domain-based analysis of the generalizability and invariance property of the SSL representations used in transfer learning classification tasks. Specifically, we show that SSL representation based models exhibit slightly better generalizability than the SL representations when the target domain is significantly different from the source domain.

  \item We study the effectiveness (generalizability) and efficiency (fine-tuning epochs to convergence) of the models that transfer learn from pretrained representations using the main SSL techniques. We show that although both the instance-based and clustering-based SSL approaches are comparable in terms of effectiveness (generalizability), the latter is more efficient.

  \item We study the generalizability of SSL-representation based models in comparison to that of the SL-representation based model by using their invariance property. We show that both the SSL and SL-based models exhibit similar invariance property when the entire input distribution goes through spatial changes such as translation, rotation, flip.

  \item We further study the generalizability of the SSL- and SL-representation based models against the variation in part of the input distribution, e.g., when the main object in the input images goes through some spatial changes or when only its background and global illumination change. We show that SSL-representation based models exhibit better generalizability as compared to the SL-representation based model.

  \item Finally, we explain better generalizability of the SSL-representation based models by examining their reasoning process in comparison to that of the SL-representation based model. By using an interpretability framework we show that SSL-representation based models' generalizability is attributed to their better invariance property that is acquired by their ability to fix a sharp focus and success in locating discriminating pixels on the input images.

        
    

\end{itemize}

The rest of the article is organized as follows. In section 2, we discuss the background and relevant literature on SSL. In section 3, we describe the experimental setting, datasets, learning techniques, interpretability framework, and study design. Results obtained from the experiments are provided in section 4 followed by a detailed analysis. Section 5 presents the conclusion of this article with an outline of future work.   

\section{Background}
\label{chap:ssl}

\begin{figure}[htb!]
\centering
\includegraphics[width=12cm]{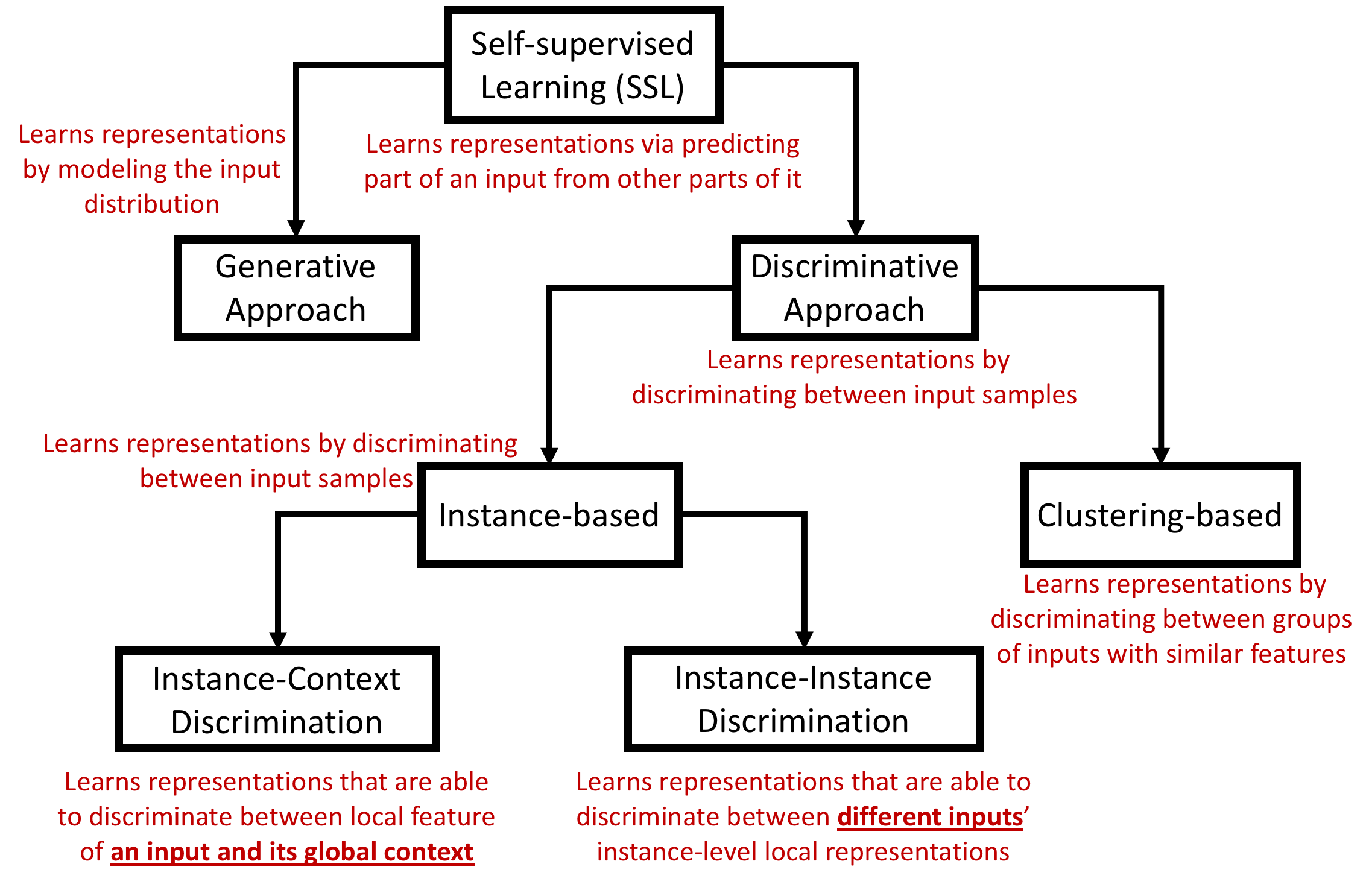}
\caption{Main approaches of SSL.}
\label{fig:ssl-techniques}
\end{figure}

The core idea behind SSL is to automatically generate supervisory signals that helps the algorithm solve a specific task. Algorithms that fall under this type of learning do not use annotated datasets for learning representations. In general, there are two approaches for creating representations from unlabeled data: generative and discriminative. In this article, we focus on the discriminative approach and discuss various techniques under two main categories as shown in fig \ref{fig:ssl-techniques}.

				
				

\subsection{Generative Approach to Self-Supervised Learning}
Generative Learning aims to learn representations by modeling the input distribution $p(x)$. This is done under the assumption that a good model of $p(x)$ contains sufficient information about the category distribution $p(y | x)$. More informally, this means that if we can learn a representation that is good enough for perfect reconstruction, then the same representation encodes sufficient information about the category of the sample, providing it with the ability to discriminate between samples. Along with this assumption, we expect the learned representations to be invariant to changes in samples irrelevant to the task. Invariance is generally achieved via regularization of latent representations (enforcing sparsity \citep{olshausen:1996} or robustness to noise \citep{bengio:2014, vincent:2008}).

Ideally, a generative model of natural images would be capable of generating images under their natural distribution. It does so by learning the latent structure of the input distribution. However, inferring latent structures, given an image, is intractable even for simple models. There has been some work to overcome this limitation, e.g., Wake-Sleep algorithm \citep{hinton:1995}, Contrastive Divergence \citep{hinton:2006}, Deep Boltzmann Machines \citep{salakhutdinov:2009}, and Variational Bayesian methods \citep{kingma:2014, rezende:2014}. These techniques use sampling to perform approximate inference. Since generative models attempt to directly model the pixel distribution, which is computationally expensive, it becomes difficult to train a CNN. Given the way these models work and their limitations, they have not been great at performing well on high-resolution natural images. This approach is only feasible for smaller datasets such MNIST handwritten digits \citep{hinton:1995, hinton:2006, salakhutdinov:2009, kingma:2014, rezende:2014}.

\subsection{Discriminative Approach to Self-Supervised Learning}
The Discriminative approach to SSL is fundamentally different from the Generative approach. Generative models learn how to reconstruct an input but this seems unnecessary in the context of tasks like classification, detection, etc. If the task is to classify categories, then image reconstruction is unnecessary and increases complexity. A discriminative approach instead attempts to learn an embedding (a feature vector for each image). In a latent space, semantically similar embeddings would be close to each other while different ones would be further apart.

The goal of the discriminative approach is not to learn the input distribution, instead it aims to learn a representation that can discriminate between input samples. It uses objective functions such as cross-entropy for learning representations that are similar to the ones used in supervised learning \citep{dosovitskiy:2015}. However, unlike in SL, these functions are used to train networks for performing pretext tasks where both inputs and labels are derived from an unlabelled dataset. A pretext task is defined as a task that is not directly useful, but is used to create good representations as a byproduct of training the model. An example of a pretext task is to spatially order patches from a static image \citep{doersch:2017} or predicting various rotated versions of an image \citep{gidaris:2018}. Representations that are learned using this approach are discriminative as well as invariant to some typical transformations. The discriminative and invariance property of the representations make them useful for vision tasks. 


The various techniques of the discriminative SSL approach can be grouped into two categories.
\begin{itemize}
     \item Instance-based: It focuses on learning how to discriminate specific instances instead of classes.
     \item Clustering-based: It is similar to how the SL approach learns to discriminate between classes/categories but in absence of labels the algorithm has to rely on the underlying structure of the dataset.
\end{itemize}

\subsection{Instance-based Approach}

This approach learns a good representation by capturing the apparent similarity among instances, instead of classes. It does so by forcing the features to be discriminative with respect to individual instances. Its focus on capturing latent information at the instance level enables this approach to create more generalizable representations for transfer learning than the SL approach \citep{chen:2020v2}. In the SL approach, visual representations are learned by minimizing intra-class variation, which is done via optimizing the cross-entropy loss between predictions and labels. The SL approach emphasizes the discriminative regions among the samples belonging to a class. However, its class-based learning results in loss of information in other regions \citep{singh:2017}. Thus the SL pretraining approach is unable to preserve unique information of the instances belonging to the same class. The SL approach is based on a strong assumption, i.e., all instances within a category should be alike in their latent space of representations. This assumption hurts the generalizability of transfer learning in two ways. First, fine-tuning models on the target dataset containing noisy labels, diverse backgrounds, and varying illumination (e.g., in camera-trap images) is less effective \citep{norouzzadeh:2018}. Second, it neglects all unique information from a single instance that could have been significant in downstream applications \citep{zhao:2021}. Also, a strong higher-level representation is not critical for transferring to downstream tasks \citep{zhao:2021}. On the other hand, the instance-based pretraining approach can preserve unique information of the instances that are similar. That is why the instance-based approach can create more generalizable representations. The two main approaches of the instance-based category are Instance-context discrimination and instance-instance discrimination.


\subsubsection{Instance-Context Discrimination}

This approach leverages the spatial structure among the local components of the image data. It learns representations that are able to discriminate between local features of an input and its global context. This is accomplished by predicting the relative spatial position of the local components. For making accurate predictions, the model needs to discriminate the global context of the input. The learning approach is designed as a supervised approach. However, unlike the supervised approach, the supervisory signals do not come from data. Instead, these signals are derived from heuristic-based pretext tasks. A pretext task is defined as a task by solving which a model is able to learn the context. This approach is inspired by self-supervised learning in the text domain where context is used as a source of supervisory signals. A text embedding model learns representations of a word by learning to predict its context, i. e., next and/or previous words \citep{mikolov:2013, pennington:2014}.  This ``context prediction'' task is a pretext for pushing the model to learn a good representation of the words in a text. In a similar vein, in the vision domain, good representations are learned by training a model to solve a pretext task. Examples of pretext tasks in vision are predicting relative positions of two patches from a sample \citep{doersch:2015}, recovering spatial relation of a static shuffled image by solving a jigsaw puzzle \citep{noroozi:2016}, predicting the degree of the rotation angle of a rotationally transformed image \citep{gidaris:2018}, colorizing grayscale images \citep{zhang:2016, larsson:2016}, correlating the egomotion (self-motion) of a vehicle between two consecutive frames \citep{agrawal:2015}, predicting pixels that will move in subsequent frames given a single frame of video \citep{pathak:2017}. For solving these pretext tasks, the model needs to learn the global relationship of the local components of the image. A good pretext task will allow the intermediate layers of a CNN to encode high-level semantic information that is beneficial to solving downstream tasks of interest (recognition, detection, etc.). 

The pretext-task based instance-context discrimination techniques used for pretraining representations are not very effective in transfer learning problems as compared to the SL-based pretrained representations \citep{gidaris:2018, kolesnikov:2019}. The poor transfer learning performance of these representations is caused by their reliance on the global context of an instance as the supervisory signal for learning its representations. The encoded context in their latent representation space hurts their generalizability. The context information of the training instances (i.e., background) is not useful for downstream tasks such as image classification. It is more important to learn the representations of the objects that need to be discriminated, not their context. In other words, effective transfer learning requires invariance in the representations such that even if the objects are transformed (e.g., translation, rotation, or change in the background), their representations remain unchanged. The main limitation of the instance-context discrimination techniques is that the representations are not invariant, but covariant \citep{misra:2020}.

\subsubsection{Instance-Instance Discrimination}

This approach identifies that the key factor for learning invariant representations, i.e., to solely learn the local representations of an instance irrespective of its global context. To achieve this, it needs to discriminate or contrast the instance from other instances. Thus, by being able to contrast the local representations of instances, this approach achieves invariance in the representation space. The self-supervisory signals in this approach are created by computing the contrastive loss between an instance and other instances. The contrastive loss function is based on the Noise Contrastive Estimation Loss (NCE Loss) \citep{gutmann:2010}.

The main idea of NCE is that it provides an efficient technique for estimating the density of a multi-class distribution by converting it into a binary distribution, which consists of the given instance (positive sample) and all other instances (negative samples) \citep{mnih:2012}. However, for optimal performance, it requires a large number of ``negative'' instances to contrast with the ``positive'' instance. For its effective use in representation learning, the instance-instance discrimination approach requires a lot of negative/dissimilar samples so that the network can learn to separate these instances in the latent manifold. If the number of negative samples is smaller, then the model will be forced to learn a trivial or collapsed solution. To overcome this problem, initial instance-instance discrimination techniques such as MoCo, PIRL, and SimCLR focused on strategies to include a large number of negative samples for contrastive learning.

The Momentum Contrast (MoCo) \citep{he:2020} poses the learning problem as a dictionary lookup problem where the dictionary is updated with a momentum averaging. It trains a visual representation encoder by matching an encoded query to a dictionary of encoded keys using a contrastive loss. The dictionary is used as a queue of data samples. The dictionary is dynamic as the keys are randomly sampled. Also, the key encoder evolves during training. For learning good representations, the dictionary needs to be large for being able to include many negative samples.

The Pretext-Invariant Representation Learning (PIRL) \citep{misra:2020} combines pretext task-based supervisory signal with contrastive learning to create invariant representations. The pretext task used is to recover the original order of randomly shuffled patches of an image. PIRL uses a siamese network where one network takes the original image as the input and the other takes the transformed version. The goal of the pretraining task is to encourage the representations of both of these images to be similar by using the contrastive loss function. Similar to MoCo, it maintains a memory bank for storing a large amount of samples. 

The simple framework for contrastive learning of visual representations (SimCLR) \citep{chen:2020v1} provides a contrastive loss based learning strategy without requiring to use memory banks (such as in MoCo) or special architectures (such as in PIRL). To achieve invariance, it employs data augmentation and optimizes the use of contrastive loss function by utilizing a large batch of negative samples. It further improves the invariance property by applying architectural/engineering tricks. For example, it increased the depth and width of the pretraining models, used increased batch size, and implemented a channel-wise attention mechanism known as the Selective Kernel for tuning the filters \citep{chen:2020v2}. The Selective Kernel technique \citep{li:2019} utilizes attention to pick filters in a convolution layer that are more suited to the scale of the object in the input. These tricks along with a unique distillation technique helped SimCLR to achieve state-of-the-art results on various benchmarks \citep{chen:2020v2}. 

The key issue with the SimCLR technique is its dependence on a very large batch size. The Barlow Twins (BT) technique \citep{zbontar:2021} overcome this issue by utilizing an alternative strategy to avoid collapsed solutions. It constructs an objective function that avoids collapsed solution by measuring the cross-correlation matrix between the outputs of two identical inputs fed with distorted views of a sample. The goal is to make this cross-correlation matrix as close to the identity matrix as possible. An identity would imply that the views are of the same image. Thus, this technique facilitates achieving invariance to distortions. The fact that BT does not require engineering tricks (e.g., careful tuning, large batch size) enables it to democratize the instance-instance discrimination approach by providing an efficient solution for creating generalizable representations.

\subsection{Clustering-based Approach}

The main issue with the instance-based approach is that pairwise comparisons are computationally expensive. To overcome this issue, the clustering-based techniques such as DeepCluster and SwAV learn representations by discriminating samples based on their cluster assignments. 

The DeepCluster \citep{caron:2018} technique jointly learns the representations and the cluster assignments of the representations. It uses the K-Means clustering algorithm to iteratively group the representations. The cluster assignments are used as supervisory signals to update the representations. This technique is improved in DeepCluster-v2 \citep{caron:2021} by using various tricks such as stronger data augmentation and explicit comparisons between K-Means centroids. However, the main limitation of DeepCluster-v2 is that it is infeasible on very large dataset.

To overcome this limitation, the Swapping Assignments between multiple Views of the same image (SwAV) \citep{caron:2021} technique uses an \textbf{online} algorithm. It utilizes contrastive learning without computing pairwise comparisons. It clusters samples while enforcing consistency between cluster assignments for different views (augmentations) of the same image. In a normal instance-based contrastive learning setting, the representations of these views are compared directly. However, SwAV uses a swapping mechanism where it predicts the code of a view from a representation of another view. SwAV not only avoids the requirement for large batch size, but also it does not require an external memory bank. Thus, it converges faster and is scalable.

\section{Method}
\label{chap:method}

To address the research questions given in the introduction, we design a set of studies. First, we describe the datasets used in the studies and motivate our choices. Then, we present various SSL techniques as well as the SL technique, and the network architectures. Finally, we present the interpretability study framework followed by a description of the studies.

\subsection{Datasets}
To address the first research question, we use two types of target datasets similar to and different from the source dataset (i.e., ImageNet) that is used to create the pretrained representations using both the SL and SSL techniques.

\begin{itemize}
    \item Target dataset similar to the source dataset
    
    \item Target dataset different from the source dataset
\end{itemize}

\begin{figure}[!htb]
\minipage{0.24\textwidth}
  \includegraphics[width=\linewidth]{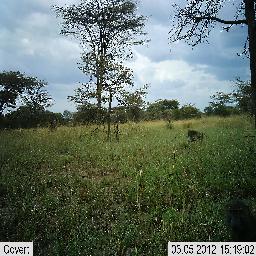}
\endminipage\hspace{1.3pt}
\minipage{0.24\textwidth}
  \includegraphics[width=\linewidth]{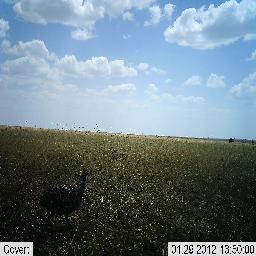}
\endminipage\hspace{1.3pt}
\minipage{0.24\textwidth}%
  \includegraphics[width=\linewidth]{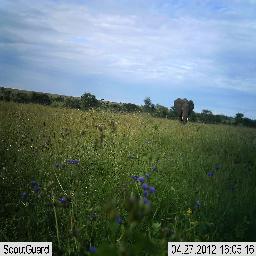}
\endminipage\hspace{1.3pt}
\minipage{0.24\textwidth}%
  \includegraphics[width=\linewidth]{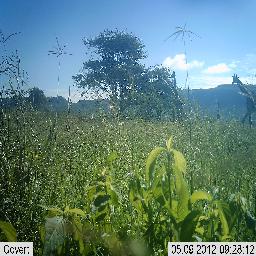}
\endminipage
\caption{Snapshot Serengeti Dataset: Prominence of background.}
\label{fig:Serengeti-Background}
\end{figure}
\begin{figure}[!htb]
\minipage{0.24\textwidth}
  \includegraphics[width=\linewidth]{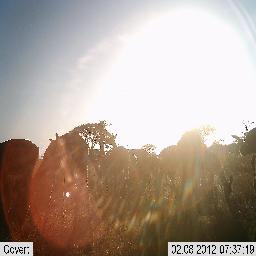}
\endminipage\hspace{1.3pt}
\minipage{0.24\textwidth}
  \includegraphics[width=\linewidth]{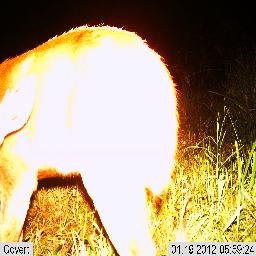}
\endminipage\hspace{1.3pt}
\minipage{0.24\textwidth}%
  \includegraphics[width=\linewidth]{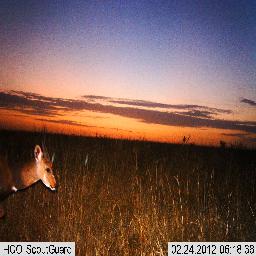}
\endminipage\hspace{1.3pt}
\minipage{0.24\textwidth}%
  \includegraphics[width=\linewidth]{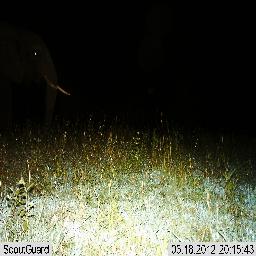}
\endminipage
\caption{Snapshot Serengeti Dataset: Changes in illumination.}
\label{fig:Serengeti-Illumination}
\end{figure}
\begin{figure}[!htb]
\minipage{0.24\textwidth}
  \includegraphics[width=\linewidth]{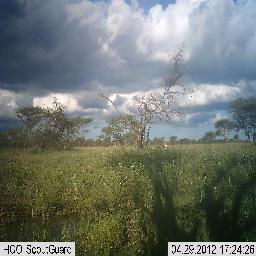}
\endminipage\hspace{1.3pt}
\minipage{0.24\textwidth}
  \includegraphics[width=\linewidth]{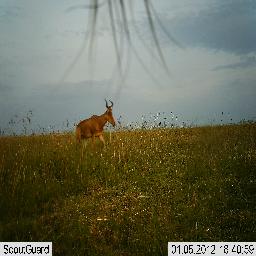}
\endminipage\hspace{1.3pt}
\minipage{0.24\textwidth}%
  \includegraphics[width=\linewidth]{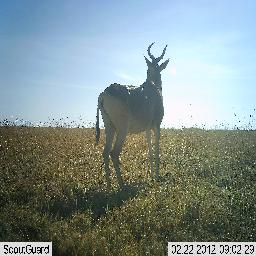}
\endminipage\hspace{1.3pt}
\minipage{0.24\textwidth}%
  \includegraphics[width=\linewidth]{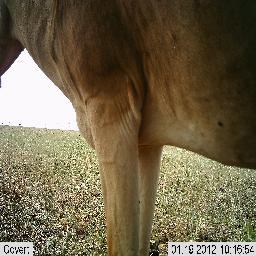}
\endminipage
\caption{Snapshot Serengeti Dataset: Scale of the Hartebeest varies.}
\label{fig:Serengeti-VaryingScale}
\end{figure}
\begin{figure}[!htb]
\minipage{0.24\textwidth}
  \includegraphics[width=\linewidth]{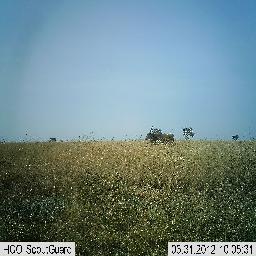}
\endminipage\hspace{1.3pt}
\minipage{0.24\textwidth}
  \includegraphics[width=\linewidth]{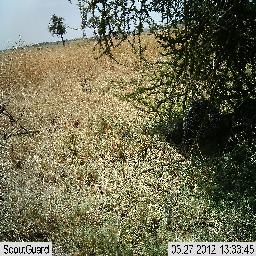}
\endminipage\hspace{1.3pt}
\minipage{0.24\textwidth}%
  \includegraphics[width=\linewidth]{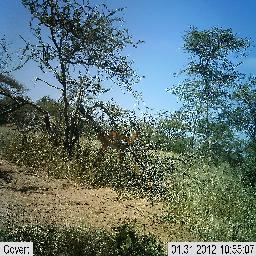}
\endminipage\hspace{1.3pt}
\minipage{0.24\textwidth}%
  \includegraphics[width=\linewidth]{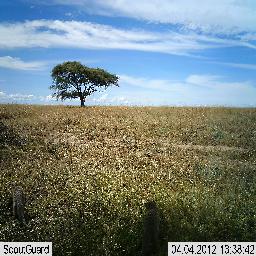}
\endminipage
\caption{Snapshot Serengeti Dataset: Objects blended in the background.}
\label{fig:Serengeti-Blended}
\end{figure}
\begin{figure}[htb!]
\centering
\includegraphics[width=10cm]{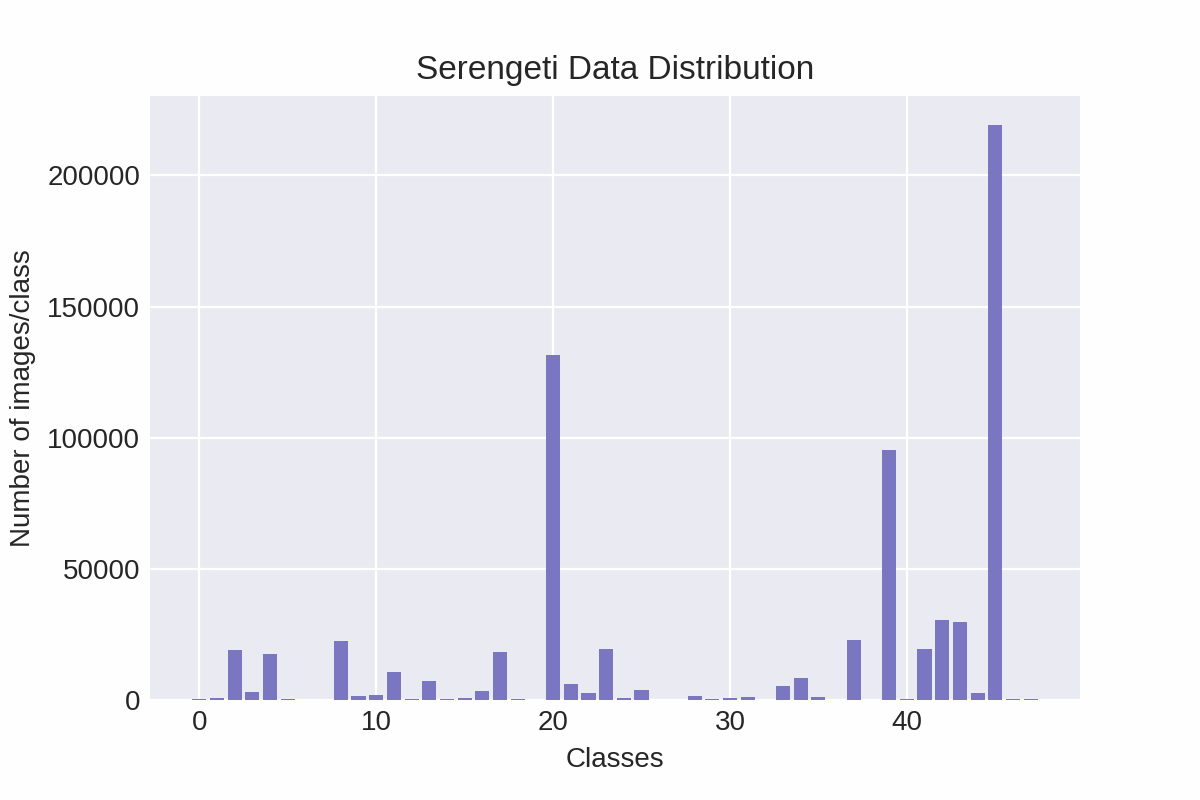}
\caption{Data distribution for the Serengeti dataset.}
\label{fig:serengeti-data}
\end{figure}

For the first type, we use the CIFAR-10 and CIFAR-100 as target datasets due to their similarity with the source dataset. For the latter type, we select a domain that is \textbf{significantly different} from the source dataset. We use the Snapshot Serengeti \citep{Swanson:2015} camera-trap dataset for this type. We argue that the camera-trap data domain is considerably dissimilar to the source domain. More specifically, camera-trap images are different from ImageNet/CIFAR like images because:

\begin{itemize}
    \item They suffer from heavily skewed distributions of classes as shown in figure \ref{fig:serengeti-data}. For example, in the Serengeti dataset, three classes occupy more than 63\% of the images. This is because some animals have a higher population and tend to move around more. Thus, they are captured more.
    
    \item The images often have noisy labels. A camera-trap captures images as a set of pictures labeled with the same category. However, the object of interest in the capture event images with the same label may not present in all images, or may be partially present as shown in figure \ref{fig:Serengeti-CaptureEvent-GazelleGrants}.
    

\begin{figure}[htb!]
\minipage{0.32\textwidth}
  \includegraphics[width=\linewidth]{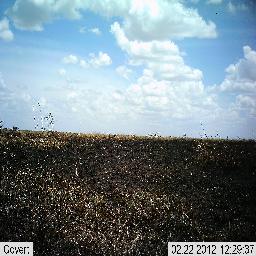}
\endminipage\hfill
\minipage{0.32\textwidth}
  \includegraphics[width=\linewidth]{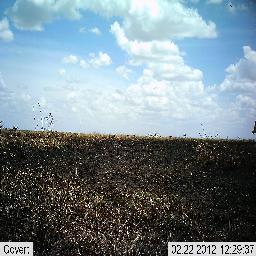}
\endminipage\hfill
\minipage{0.32\textwidth}%
  \includegraphics[width=\linewidth]{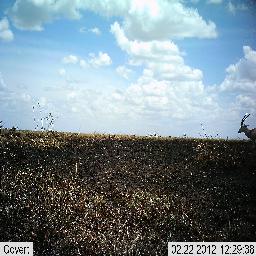}
\endminipage
\caption{Snapshot Serengeti Dataset: Capture Event of 3 images with the same label ``Gazelle Grants''. Left: no animal on the image. Middle: animal is partially visible. Right: Animal is mostly visible.}
\label{fig:Serengeti-CaptureEvent-GazelleGrants}
\end{figure}

    
    \item The presence of background in the images is more prominent as shown in figure \ref{fig:Serengeti-Background}. 
    

    \item The global illumination varies significantly in the images as shown in figure \ref{fig:Serengeti-Illumination}.

     
     \item Scale of the objects vary in the images as shown in figure \ref{fig:Serengeti-VaryingScale}

     \item In some images, objects are blended in the background, hence difficult to discriminate, as shown in figure \ref{fig:Serengeti-Blended}.

\end{itemize}

 \begin{figure}[!htb]
\minipage{0.24\textwidth}
  \includegraphics[width=\linewidth]{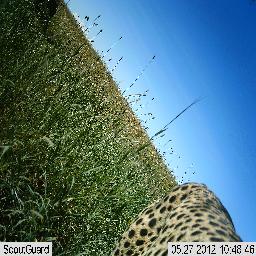}
\endminipage\hspace{1.3pt}
\minipage{0.24\textwidth}
  \includegraphics[width=\linewidth]{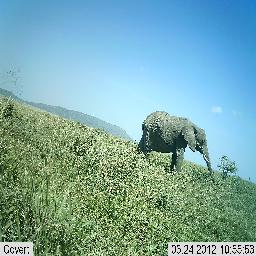}
\endminipage\hspace{1.3pt}
\minipage{0.24\textwidth}%
  \includegraphics[width=\linewidth]{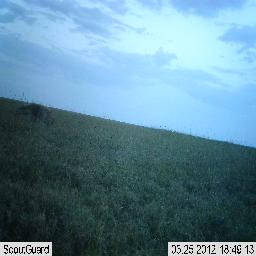}
\endminipage\hspace{1.3pt}
\minipage{0.24\textwidth}%
  \includegraphics[width=\linewidth]{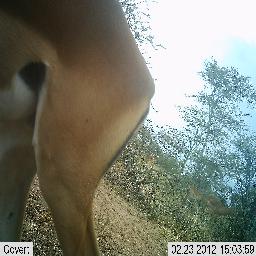}
\endminipage
\caption{Snapshot Serengeti Dataset: Objects are rotated.}
\label{fig:Serengeti-Rotated}
\end{figure}





Below we describe the datasets used in this research.

\paragraph{Snapshot Serengeti}
We choose the Snapshot Serengeti dataset \citep{Swanson:2015} because it represents a real-world natural image dataset. The domain of this dataset is significantly different from the ImageNet domain. The class distribution is imbalanced and the dataset has a lot of noise (in the form of images labeled with a category even when they are empty, contains occlusions, and varied illumination). The Snapshot Serengeti Project is the world's largest camera-trap project. It has 225 camera traps that continuously capture images. The public dataset consists of millions of images. We follow the preprocessing strategy outlined in \citep{norouzzadeh:2018} to curate a dataset with 757,000 images distributed over 48 classes. We use their gold-standard dataset as a validation set, which is annotated by experts. On the other hand, the training dataset is annotated by volunteers. We employ the data augmentation described in \citep{norouzzadeh:2018}.

 \begin{figure}[!htb]
\minipage{0.55\textwidth}
  \includegraphics[width=\linewidth]{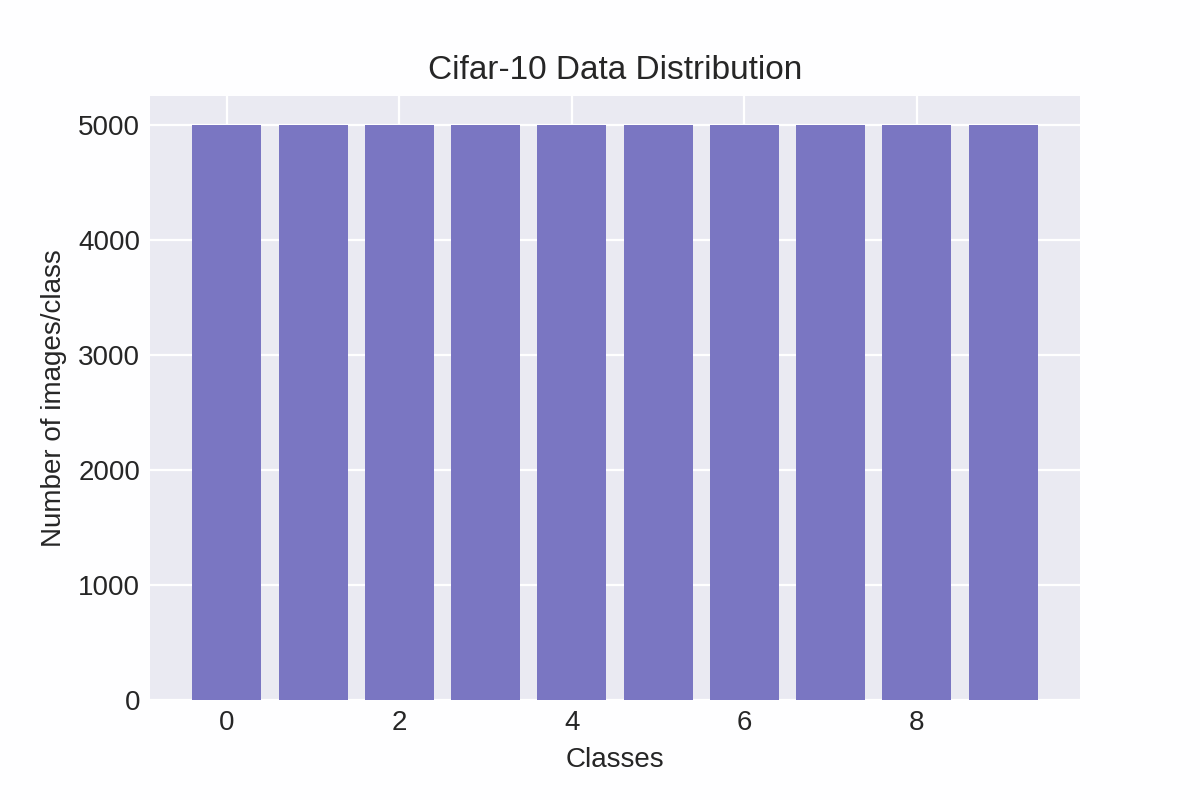}
\endminipage\hspace{0.0pt}
\minipage{0.55\textwidth}
  \includegraphics[width=\linewidth]{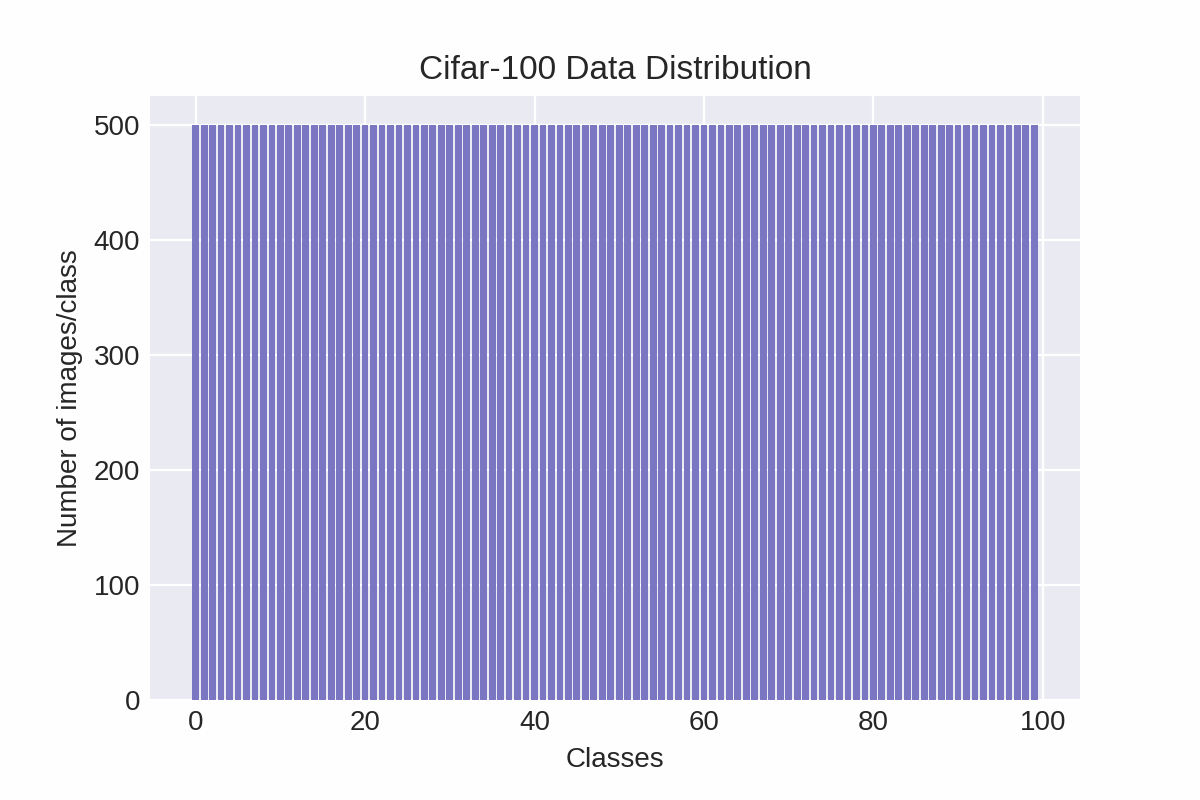}
\endminipage\hspace{0.0pt}
\caption{Data distribution for the CIFAR-10 and CIFAR-100 dataset.}
\label{fig:Serengeti-Rotated}
\end{figure}

\paragraph{CIFAR 10 \& 100}
Finally, we use two ImageNet-type dataset, i.e., the CIFAR datasets \citep{Krizhevsky:2010}. These datasets exhibit similarity to ImageNet dataset in terms of the distribution of the categories and nature of the images. 


%
%
%

\subsection{SSL Techniques}
All representations (both self-supervised and supervised) are pretrained on the same source dataset, i.e., the ImageNet dataset, and then fine-tuned for downstream classification task using the above-mentioned target datasets.

We investigate two types of SSL techniques for pretraining representations.

\begin{itemize}
    \item Instance-based: SimCLRv2 and Barlow Twins
    \item Clustering-based: DeepClusterv2 and SwAV
\end{itemize}

\subsubsection{SSL: Instance-based}
\paragraph{SimCLRv2}
We use SimCLRv2, which is one of the most effective instance-based SSL technique. The SimCLRv2 representations are pretrained on ImageNet by using a larger batch size of 4096. For our experiments, we use SimCLRv2 pretrained on the ResNet model with varying architecture and optimization technique. 

\begin{itemize}
    \item Architecture Depth: 50 and 152.
    \item Architecture Width: 1x and 2x.
    \item Filter Calibration: With and without Selective Kernel \citep{li:2019}
\end{itemize}

\paragraph{Barlow Twins}
The second instance-based technique we select is the Barlow Twins. It uses a smaller batch size of 2048 for pretraining on ImageNet. Pretraining is done over 1000 epochs. The Barlow Twins is very efficient in terms of convergence, batch size, and its ability to form good representations without much tuning. 


\subsubsection{SSL: Clustering-based}
\paragraph{SwAV}
We select the clustering-based Swapping Assignments between multiple Views (SwAV) technique because of its effectiveness. Unlike SimCLRv2, SwAV can be pretrained on smaller batch and is proven to converge much faster. We utilize two variants of ResNet-50 pretrained with SwAV (1x and 2x width). The ResNet-50 1x variant is pretrained on ImageNet for 800 epochs with a batch size of 4096, and the 2x variant is pretrained for 400 epochs with a batch size of 4096.


\paragraph{DeepClusterv2}
We use another clustering-based technique namely the DeepClusterv2. The model is pretrained for 800 epochs with a batch size of 4096. We only use the ReNet-50 pretrained model.


\subsubsection{SL}

We use the dominant SL approach for pretraining that is based on the cross-entropy loss function \citep{kolesnikov:2019}. We experiment with two variants of ResNet (50 and 152) pretrained on ImageNet with the SL approach.

\subsection{Interpretability Study Framework }

Apart from studying the domain-based efficacy of the SSL techniques, we want to \textbf{explain} the generalizability as well as understand the reasoning process of the pretrained SSL-representation based fine-tuned models. For this, we design an interpretability study framework. It addresses the following questions:

\begin{itemize}
    \item How do we explain the generalization capability of the SSL-based models?
    
    \item How do the SSL-based models reason about the semantic identity of the data?
\end{itemize}

For addressing the first question, we study the invariance property of the models. The latter question is addressed by using filter visualization technique.


\subsubsection{Generalizability Analysis via Invariance Property}


The pretrained representations are generalizable if after fine-tuning the model performs effectively on the target dataset. For explaining the generalizability, we study the invariance property of the fine-tuned model. Since the probability distribution of data (features, and semantic identity) is invariant to transformations (e.g., translation, rotation, reflection, scale change), we expect the visual representations learned by a DL model to be invariant as well. Invariance also refers to a model's robustness in predictions with respect to the changes in the background and global illumination. Invariance is attributed to the representations learned by the final layer of a model. A model's final layer representations are invariant if its predictions remain unchanged even when the input or the background is transformed.


The framework investigates the invariance of SSL-based models' predictions against various types of transformations of the input images. The transformations are of three types: the objects that the fine-tuned model attempts to classify are spatially transformed, the image background is varied, and the variation in the global illumination. The dominant SSL techniques such as PIRL \citep{misra:2020}, SimCLR \citep{chen:2020v1, chen:2020v2}, MoCo \citep{he:2020}, SwAV \citep{caron:2021}, and Barlow Twins \citep{zbontar:2021} exhibit some degree of invariance that results in their improved generalization capability. However, it is not clear whether the invariance achieved via the SSL-based models is different from that obtained from the SL-based models. Our framework is used to study the invariance of SSL-based models in two ways: (i) by passing spatially transformed images through a fine-tuned model and evaluating its performance through prediction probabilities, (ii) by attribution study (discussed next) to observe how a model processes the variation in location of the main object, its scale, background and global illumination of an input in its final convolutional layer. We describe these two approaches in study 3 and study 4. 

\subsection{Generalizability Analysis via Attribution}

To further study the invariance property as well as understanding the SSL-based model's reasoning process, the framework uses the attribution technique. Attribution study shows what part of an input is processed by a network for making predictions \citep{selvaraju:2017, wang:2020}.

\subsubsection{Attribution using Saliency Map for Original Image}
The attribution study is done by using a saliency map for the original image. The saliency maps or heat maps are created by computing the gradient of the output (neuron/channel/layer) with respect to the input while holding the weights fixed. This determines which input elements (e.g., which pixels in case of an input image) need to be changed the least to influence the output the most. A saliency map is based on the heat map of class activation that is superimposed on the original input image. A class activation heat map provides scores associated with a specific output class on a two-dimensional grid. The scores are computed for every location in any input image. These scores indicate how important each location is with respect to the class used for analysis. The saliency maps will be used to understand why does a model decide a particular object to be present in the input image. It will also show where the model thinks that the object of interest is located. For creating saliency maps, we use the Score-CAM technique \citep{wang:2020}.

\subsection{Study Design}

To address our research questions in the introduction, we design the following studies.

\subsubsection{Study 1 (to address RQ1)}
Fine-tune (ImageNet-based) SL pretrained CNNs using the following datasets: CIFAR-10, CIFAR-100, and Snapshot Serengeti. The goal of this study is to setup a benchmark for the SL approach to compare it with the SSL techniques. 

\subsubsection{Study 2 (to address RQ1 \& RQ2)}
Fine-tune (ImageNet-based) SSL pretrained CNNs using the following datasets:  CIFAR-10, CIFAR-100, and Snapshot Serengeti. The pretrained SSL representations are based on two main SSL approaches: instance-based (techniques are SimCLRv2 and Barlow Twins), and clustering-based (techniques are DeepClusterv2 and SwAV). The goal of this study is two-fold. First, address RQ1 by comparing the SSL-based models with the SL-based models with domain emphasis. Second, results obtained from this study will help to address RQ2, i.e., it will offer insights into whether the two main SSL approaches exhibit domain-based variance in their performance.

\subsubsection{Study 3 (to address RQ3)} 

We intend to determine whether the SSL-based models fine-tuned with the domain data exhibit prediction-invariance with respect to various transformations to the input images such as translation (vertical and horizontal), rotation, flip, and scale change, as well as change in the background and global illumination in the images. The invariance symmetry of a model is particularly useful for the camera-trap domain. We compare the invariance property of the SSL and SL approaches to understand if there exists any variation.

We select an efficient and effective SSL technique on the Serengeti camera-trap dataset for comparison, i.e., BarlowTwins \citep{zbontar:2021} pretrained ResNet-50 model. We compare its invariance property with the SL pretrained ResNet-50 model.



For spatial transformations, we select ten images from the following four categories. We select these categories based on how the SSL and SL models perform. Instead of randomly selecting images from the entire dataset for the prediction-invariance analysis, we first select a class on which a model performs well and a class on which the model performs poorly. Then, we randomly sample a set of images from these two classes. Since the dataset is heavily skewed, we use precision and recall as suitable performance metrics for model selection. We select two categories for the SSL model and two categories for the SL model.


\begin{itemize}
    \item Category 1 (Zebra): High precision/recall for both SSL and SL.
    \item Category 2 (Guinea Fowl): High precision/recall for SSL.
    \item Category 3 (Ostrich): High precision/recall for SL.
    \item Category 4 (Lion Male): Low precision/recall for both SSL and SL.
\end{itemize}

\begin{figure}[h!]
\centering
\includegraphics[width=12cm]{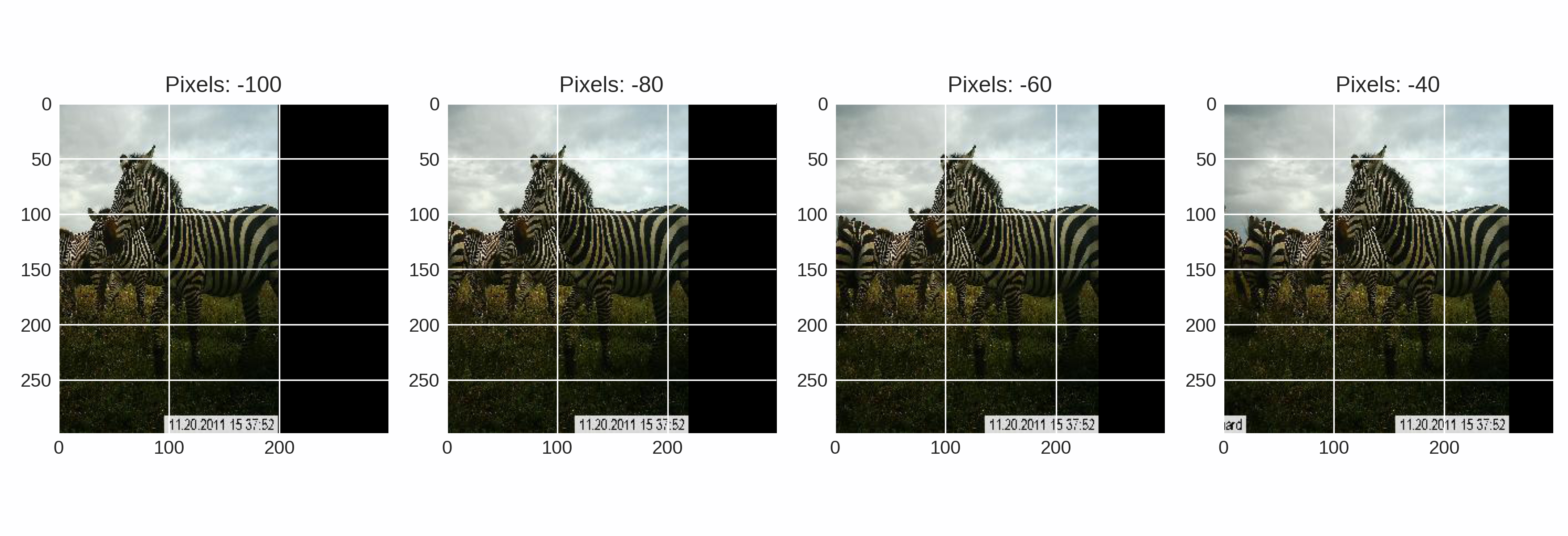}
\caption{Example of horizontal translation on a zebra image.}
\label{fig:z-ht}
\end{figure}

\begin{figure}[h!]
\centering
\includegraphics[width=12cm]{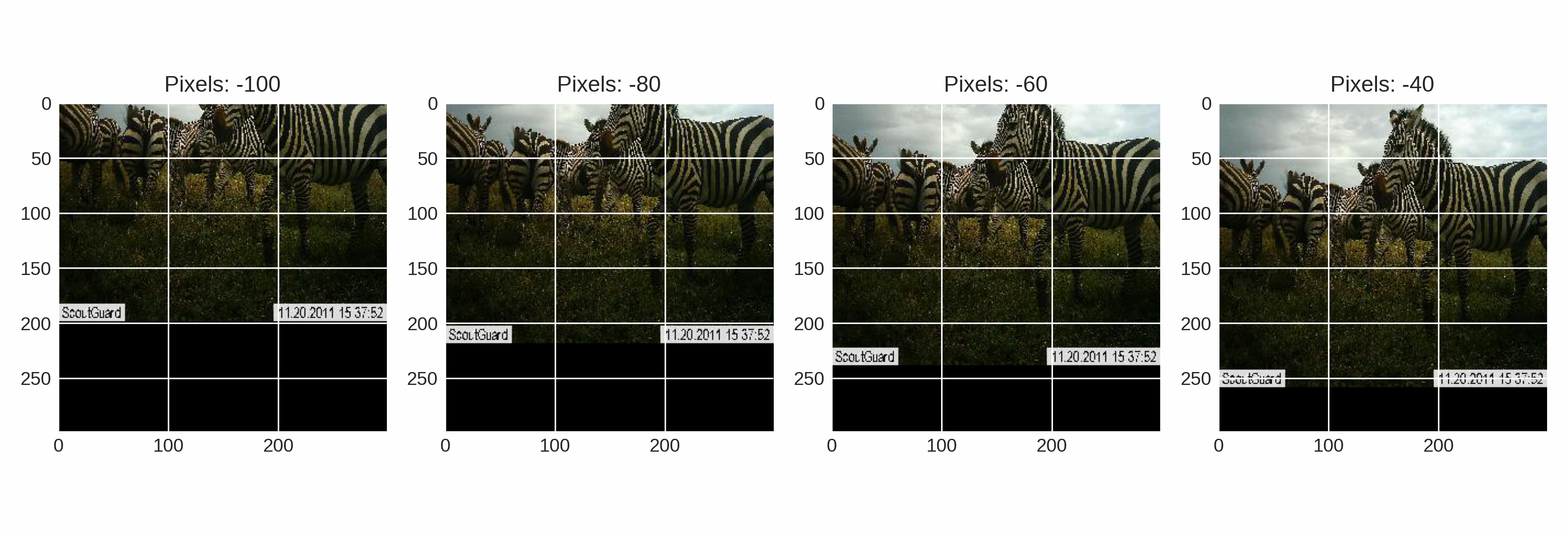}
\caption{Example of vertical translation on a zebra image.}
\label{fig:z-vt}
\end{figure}

\begin{figure}[h!]
\centering
\includegraphics[width=12cm]{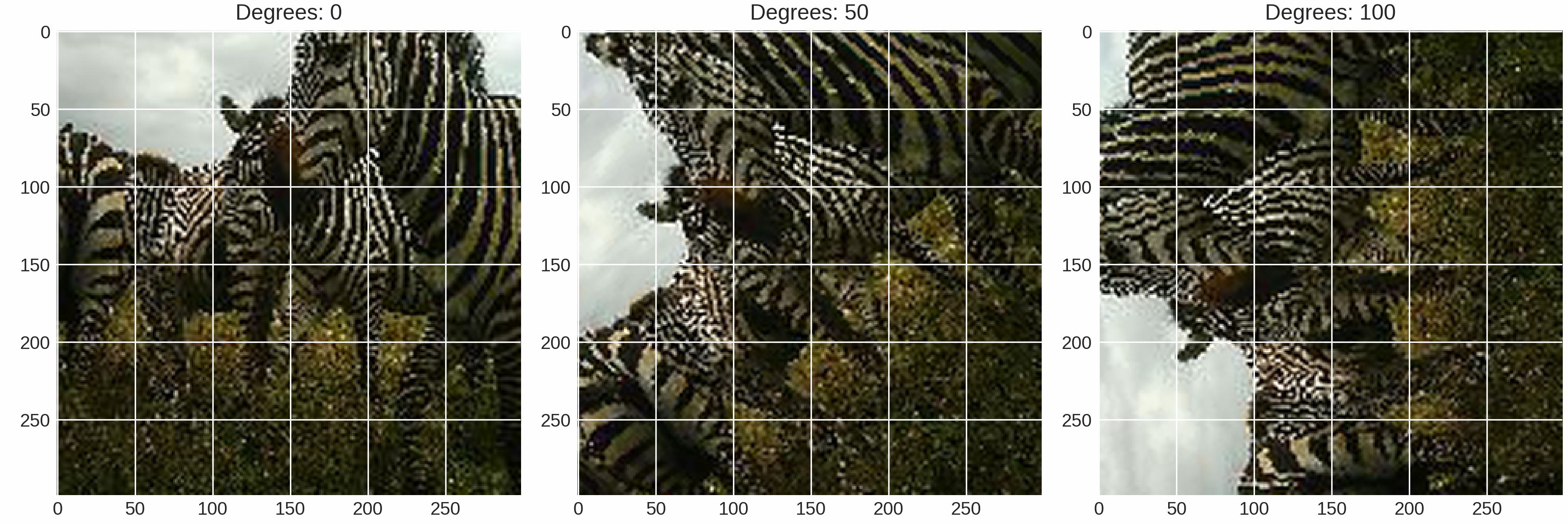}
\caption{Example of rotation on a zebra image.}
\label{fig:z-r}
\end{figure}

\begin{figure}[h!]
\centering
\includegraphics[width=10cm]{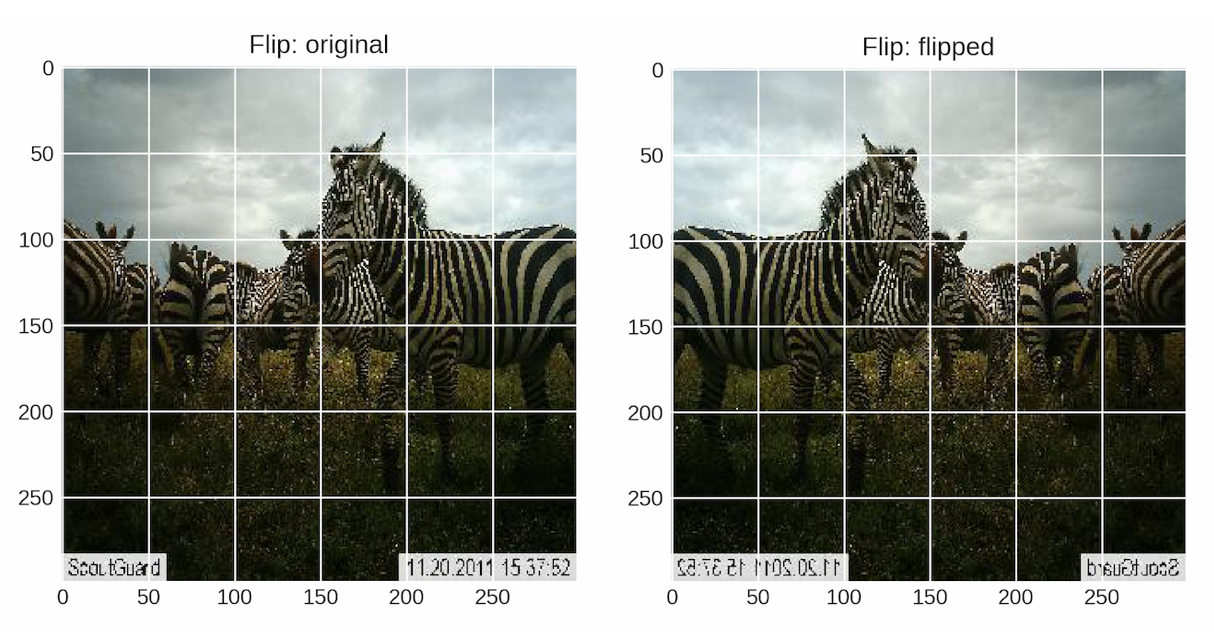}
\caption{Example of flip on a zebra image.}
\label{fig:z-f}
\end{figure}

\begin{figure}[h!]
\centering
\includegraphics[width=12cm]{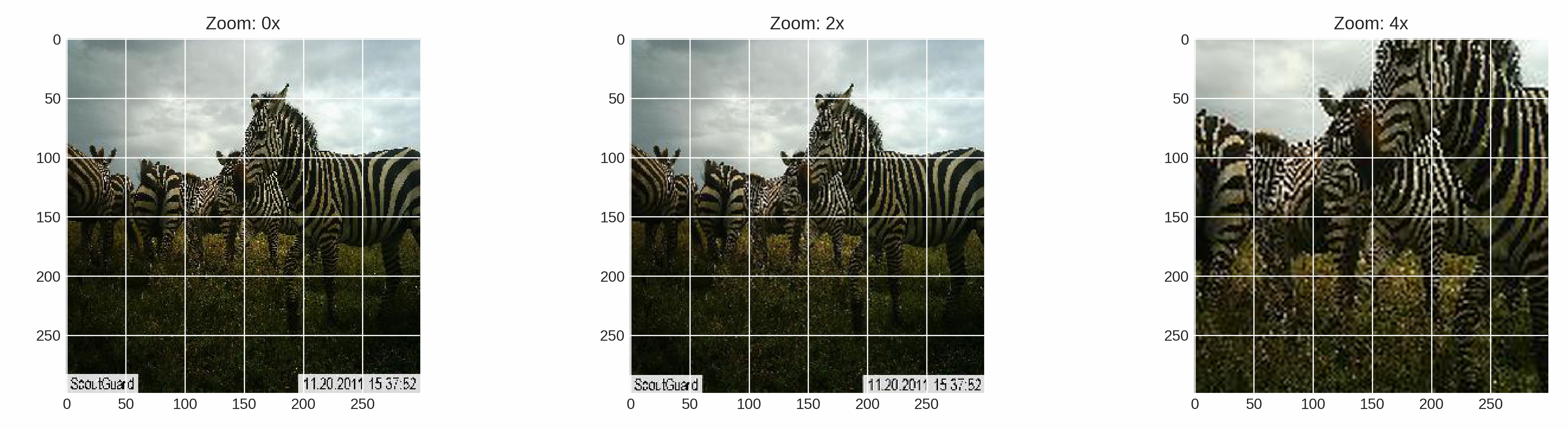}
\caption{Example of scale-zoom on a zebra image.}
\label{fig:z-s}
\end{figure}

For computing the prediction invariance, following transformations are used:
\begin{itemize}
    \item Horizontal Translation (pixels): [-100, -75, -50, -25, 0, 25, 50, 75, 100] shown in figure \ref{fig:z-ht}
    \item Vertical Translation (pixels): [-100, -75, -50, -25, 0, 25, 50, 75, 100] shown in figure \ref{fig:z-vt}
    \item Rotation (degrees): [0, 50, 100, 150, 200, 250, 300, 350] shown in figure \ref{fig:z-r}
    \item Flip: [left/right] figure \ref{fig:z-f}
    \item Scale (zoom): [0x , 2x, 4x, 6x, 8x, 10x] shown in figure \ref{fig:z-s}
\end{itemize}

For each transformation per category, we compute the average confidence/prediction probabilities for ten images. Then, we plot the average confidence against the transformations similar to \citep{zeiler:2014}. The plots provide insights into the invariance properties of the SSL and SL-based models.

\subsubsection{Study 4 (to address RQ3 and RQ4)}


The goal of this study is to analyze invariance property and interpret the decision-making process of the CNN models fine-tuned using both SSL and SL pretrained representations. We attempt to understand this process via the attribution technique. We use the same models from study 3 (i.e., ResNet-50 for BarlowTwins \citep{zbontar:2021} and SL). The  attribution technique is described in the interpretability study framework. The attribution analysis is done by selecting a fixed DL model fine-tuned on a target dataset with pretrained SSL and SL representations.






\paragraph{Attribution} To create the saliency maps for both SSL and SL models, we select images with naturally occurring variation. We use the following criteria.

\begin{itemize}

\item Images with varying translation

\item Images with various rotation

\item Images with varying scale

\item Images with varying background

\item Images with varying illumination

\item Images that are not easily discernible (main object is blended with the background or appears ambiguous)

\end{itemize}



\section{Results \& Analysis}

In this section, we describe the experiments for the four studies, discuss obtained results, followed by an analysis of the findings. 

For the studies, we conduct two types of experiments: fine-tuning and training. For the fine-tuning experiments, a pretrained model is obtained (SSL and SL), which is fine-tuned using the target data. The training experiments are done by randomly initializing the weights of a model, then training with the target data.


\subsection{Experimental Setting}
All experiments are done using the ResNet model \citep{he:2016}. Three ResNet architectures of varying capacity (varying width and depth) are used: shallow-narrow architecture (ResNet-50 1x), shallow-wide architecture (ResNet-50 2x), and deep-narrow architecture (ResNet-152 1x). SSL-based pretrained models of varying capacity are available only for SimCLR and SwAV. SimCLR provides three variants of the pretrained ResNet architectures as well as their Selective Kernel optimized versions \citep{simclr}. For the SwAV technique, available pretrained models are ResNet-50 1x and ResNet-50 2x \citep{swav}. For the Barlow Twins and DeepCluster models, the only available pretrained model is ResNet-50 1x \citep{bt, swav}. For all models, following hyperparameters are used for fine-tuning and training. 


\begin{itemize}
    \item Optimizer: Adam \citep{kingma:2017} 
    \item Learning Rate: 1e-5
    \item Weight Decay: 5e-4
    \item Batch Size: 128
    \item Epochs: 25 
\end{itemize}

We selected the learning rate based on the standard practice of utilizing a small learning rate during the fine-tuning phase \citep{mahajan:2018, keras-finetuning, cs231n}. For determining the weight decay, we experimented with various values. We noticed that the variation did not make a significant difference. We chose the one that seemed to work best. The choice of the batch size is constrained by the memory on our GPUs. We fine-tuned the models for 25 epochs, because all models converged within this number of epochs. 


The experiments are done on two Tesla V100 GPUs. Each model is trained in parallel on these two GPUs. The training time per-epoch for the large-scale dataset, i.e., Serengeti is 120 minutes on average (there is variance depending on the number of parameters in the model). The CIFAR experiments were conducted on Google Colab. Only 1 GPU is used for Colab experiments since the datasets were smaller in size and did not require big compute or storage.

\paragraph{SSL Pretrained Models}  We obtained the \textbf{SimCLRv2} pretrained models from \citep{simclr}. The batch size and pretraining epochs of these models vary \citep{chen:2020v2, simclr}. The \textbf{SwAV} pretrained models are taken from \citep{swav}. We used two variants of ResNet-50 SwAV. The 1x variant was pretrained for 800 epochs, whereas the 2x variant was pretrained for 400 epochs. Both used a batch size of 4096. The \textbf{Deep Cluster v2} pretrained model is obtained from \citep{swav}.We used the ResNet-50 1x model which was pretrained with a batch size of 4096 for 800 epochs. We got the \textbf{Barlow Twins} pretrained model from \citep{bt}. The ResNet-50 1x model is used, which was pretrained for 1000 epochs with a batch size of 2048.




\subsection{Study 1}
In this study we fine-tune ResNet models that are pretrained using the SL approach on the ImageNet dataset. The fine-tuning performance results obtained from three target datasets are given below.


\paragraph{CIFAR}
Tables \ref{table:SL-CIFAR10} and \ref{table:SL-CIFAR100} show the fine-tuning results on the two CIFAR datasets. We only use the ResNet-50 model for these experiments as the input data is low-resolution, small in size, and uniformly distributed. The fine-tuned models achieve 95.90 test accuracy on the CIFAR-10 dataset, and 81.41 test accuracy on the CIFAR-100 dataset. CIFAR-100 is inherently difficult due to the fact that there are only 500 images per category and it is a 100-class classification problem.

\begin{table}[!htb]
\scalebox{0.95}{
\begin{tabular}{ccccc}
\hline
\multicolumn{5}{c}{Supervised: CIFAR-10}                    
\\ \hline
Architecture  							& Param (M) 		& Epoch 				& Train Acc (\%) 		& Test Acc(\%) 			\\ \hline
\textbf{ResNet-50-1x} (SL) 				& 24        			& 21 					& 99.88                   		& 95.90            			\\
ResNet-50-1x  (random)  					& 24  			& 300 				& 99.30               		& 92.70                   	 	\\ \hline
\end{tabular}}
\caption{CIFAR-10: Model fine-tuned with SL representations vs. model trained with random weights.}
\label{table:SL-CIFAR10}
\end{table}

\begin{table}[!htb]
\scalebox{0.95}{
\begin{tabular}{ccccc}
\hline
\multicolumn{5}{c}{Supervised: CIFAR-100}                    
\\ \hline
Architecture  						& Param (M) 		& Epoch 				& Train Acc (\%) 		& Test Acc (\%) 		\\ \hline
\textbf{ResNet-50-1x (SL)}  			& 24        			& 25 					& 99.94               		& 81.41          		         \\
ResNet-50-1x  (random)  				& 24  			& 350 				& 99.40		              	& 65.60		              	 \\ \hline
\end{tabular}}
\caption{CIFAR-100: Model fine-tuned with SL representations vs. model trained with random weights.}
\label{table:SL-CIFAR100}
\end{table}

To understand the generalizability of the SL-pretrained model, we train the same model architecture using \textbf{randomly initialized weights} on the target data. The results show that the generalizability of the model, which is fine-tuned with SL-initialized weights, is \textbf{significantly better} than that of the model trained with randomly initialized weights. Also, we see that the SL-pretrained models converge to their optimal performance approximately 14 times faster (CIFAR-10 requires 21 epochs, and CIFAR-100 requires 25 epochs).  Thus, the SL-pretrained representations transfer more effectively and efficiently on the CIFAR datasets, the domain of which is similar to the source domain of ImageNet.

\paragraph{Snapshot Serengeti}
Table \ref{table:SLvsSOTA-Serengeti} presents the fine-tuning results for the Serengeti dataset. Since this dataset is very large, we train two ResNet architectures, one shallow (ResNet-50-1x) and one deep (and ResNet-152-1x). We want to determine whether the generalizability of the SL representations is related to the model capacity. Results show that the generalizability of the fine-tuned model increases with the increase in the model's capacity (i.e., depth). In other words, when the representations are created using a deeper architecture, they become more generalizable.

\begin{table}[!htb]

\begin{centering}
\begin{tabular}{ccccc}
\hline
\multicolumn{5}{c}{Supervised: Snapshot Serengeti}                    											\\ \hline
Architecture  					& Param (M) 		& Epoch		& Top-1 Acc (\%) 		& Top-5 Acc (\%) 	\\ \hline
\hline
\multicolumn{5}{c}{SL-Pretrained Representations + Fine-tuning}   \\
\hline
\hline
ResNet-50-1x  					& 24        			& 18			& 91.93               		& 98.47              	 \\
ResNet-152-1x 					& 58        			& 16			& 92.89               		& 98.57              	 \\

\hline
\multicolumn{5}{c}{Random Weight Initialization}   \\
\hline
\hline

\textbf{ResNet-152-1x} 		 	& 58  			& 52			& 93.62               		& 98.58              	 \\ \hline
\end{tabular}
\caption{Snapshot Serengeti: Model fine-tuned with SL representations vs. model trained with random weights.}
\label{table:SLvsSOTA-Serengeti}
\end{centering}
\end{table}

To further understand the SL-pretrained model's generalizability, we train a model using randomly initialized weights on the target data. We use the best architecture from the fine-tuned experiment (i.e., ResNet-152-1x) for a fair comparison. The results show that the model fine-tuned with SL-initialized weights is less generalizable as compared to the model trained with randomly initialized weights. Our observation is consistent with \citep{norouzzadeh:2018}. However, the efficiency of the random weight based training and SL-pretrained weight based fine-tuning varies significantly. While the SL-based model requires only 16 epochs to converge to its optimal performance, the random weight based model is required to be trained for 52 epochs.

A comparison of generalizability between the CIFAR target and the camera-trap target (i.e., Serengeti) indicates that the SL-based representations exhibit better generalizability when the target domain is similar to the source domain. Its generalizability decreases when the domain of the target data is significantly different. \textbf{This observation partially addresses our RQ1}.

\subsection{Study 2}

The study 2 examines the performance of the fine-tuned ResNet models pretrained on ImageNet using various \textbf{SSL techniques}. The results from three target datasets are based on the pretrained representations obtained from three varying architectures. In study 1, we used both shallow and deep SL-pretrained models. Similarly in study 2, we use both shallow and deep SSL-pretrained models. In addition to this, we include wide SSL-pretrained models. It has been shown that the wider models (e.g., SimCLR \citep{chen:2020v2} and SwAV \citep{caron:2021}) improve generalizability. Various architectures and target datasets used for each architecture choice are shown below.

\begin{itemize}

\item Shallow and Narrow: ResNet-50 1x (used for fine-tuning with CIFAR and Serengeti datasets)

\item Shallow and Wide: ResNet-50 2x (used for fine-tuning with CIFAR and Serengeti datasets)

\item Deep and Narrow: ResNet-150 1x (used for fine-tuning with Serengeti dataset)

\end{itemize}


\begin{table}[!htb]
\begin{centering}
\begin{tabular}{ccccc}
\hline
\multicolumn{5}{c}{Self-Supervised : CIFAR-10}                                  					 			\\ \hline
Architecture     				& Param (M) 		& Epoch		& Train Acc (\%) & Test Acc (\%) 		\\ \hline

\hline
\multicolumn{5}{c}{Shallow \& Narrow Architecture (ResNet-50 1x)}   \\
\hline
\hline

SimCLR		     			& 24        			& 21			& 98.43               & 92.37               		\\
SimCLR (SK)	 			& 34        			& 25 			& 98.62               & 94.50               		\\
SwAV		 			& 24        			& 19 			& 98.15               & 95.46               		\\
DeepCluster	  			& 24        			& 20 			& 99.95               & 95.14               		\\ 
Barlow Twins	 			& 24        			& 21 			& 99.92               & 95.54               		\\

\hline
\hline
\multicolumn{5}{c}{Shallow \& Wide Architecture (ResNet-50 2x)}   \\
\hline
\hline

SimCLR		     			& 94        			& 19			& 98.21               & 93.86              		 \\
SimCLR (SK)	 			& 140       			& 25 			& 98.50               & 95.69               		\\
\textbf{SwAV}			 	& 94        			& 25 			& 99.83               & 95.96               		\\

\hline
\end{tabular}
\caption{CIFAR-10: Model fine-tuned with SSL representations.}
\label{table:SSL-CIFAR10}
\end{centering}
\end{table}

\paragraph{CIFAR}
The results for the two CIFAR datasets are shown in tables \ref{table:SSL-CIFAR10} and \ref{table:SSL-CIFAR100}. CIFAR-10 is a relatively easier task, hence all models exhibit good generalizability. We make two key observations. First, the SSL-based shallow-narrow model architecture (24M parameters) exhibits similar performance (top-1 validation accuracy) observed in the SL-based model with the same capacity (SSL Barlow Twins = 95.54 vs. SL = 95.90). Second, increasing the model capacity makes only minor improvement on this dataset. For example, the wider SwAV-based model achieves the top test accuracy, i.e., 95.96, which is slightly higher than the top-1 accuracy (95.46) of its shallow and narrow version. All SSL models are equally efficient, i.e., the number epochs for convergence does not vary much.

\begin{table}[!htb]
\begin{centering}
\begin{tabular}{ccccc}
\hline
\multicolumn{5}{c}{Self-Supervised : CIFAR-100}                                   					 \\ \hline
Architecture     				& Param (M) 		& Epoch			& Train Acc (\%) 		& Test Acc (\%)   \\ 
\hline

\hline
\multicolumn{5}{c}{Shallow \& Narrow Architecture (ResNet-50 1x)}   \\
\hline
\hline

SimCLR		    			& 24        			& 25 				& 98.91               		& 76.26               \\
SimCLR (SK)	  			& 34        			& 17 				& 97.31               		& 78.27               \\
SwAV		 			& 24           		& 22 				& 94.82               		& 76.98               \\
DeepCluster	 			& 24        			& 24 				& 96.51               		& 78.68               \\
Barlow Twins	 			& 24        			& 17 				& 96.75               		& 78.36               \\

\hline
\hline
\multicolumn{5}{c}{Shallow \& Wide Architecture (ResNet-50 2x)}   \\
\hline
\hline

SimCLR		     			& 94        			& 14 				& 96.91               		& 79.57               \\
\textbf{SimCLR (SK)}  		& 140      		        & 18 				& 96.98               		& 80.90               \\
SwAV		 			& 94        			& 25 				& 99.81               		& 79.22               \\

\hline
\end{tabular}
\caption{CIFAR-100: Model fine-tuned with SSL representations.}
\label{table:SSL-CIFAR100}
\end{centering}
\end{table}

For the CIFAR-100 dataset, the SSL-based shallow and narrow model architecture (24M parameters) performs poorly than the SL-based model with the same capacity (DeepCluster SSL = 78.68 vs. SL = 81.41). This performance gap is shown to be reduced by increasing model capacity. The wider SimCLR model with Selective Kernel optimization gets very close to the SL-based model's top-1 accuracy within only 17 epochs of fine-tuning (Wide SimCLR SSL with Selective Kernel = 80.90 vs. SL = 81.41). For both datasets, SSL representations transfer effectively to tasks that are similar to the original domain. \textbf{This observation partially addresses our RQ1}.





%
%
%

\begin{table}[!htb]
\scalebox{0.95}{
\begin{tabular}{ccccc}
\hline
\multicolumn{5}{c}{Self-Supervised : Snapshot Serengeti}                                   									\\ \hline

Architecture     							& Param (M) 		& Epoch		& Top-1 Acc (\%) 	& Top-5 Acc (\%) 	\\ \hline

\hline
\multicolumn{5}{c}{Shallow \& Narrow Architecture (ResNet-50 1x)}   \\
\hline
\hline

SimCLR		    						& 24        			& 20			&  90.62 	                 & 97.10              	 \\
\textbf{SimCLR (SK)	}  					& 34        			& 12			&  \textbf{93.87}        & 98.42               	 \\
SwAV								& 24        			& 6			&  92.14               	& 98.75              	 \\
DeepCluster	  						& 24        			& 10 			&  92.95               	& 98.94               	 \\
Barlow Twins	 						& 24        			& 11			&  92.79               	& 98.69               	 \\
Random Weight						& 24	  			& 31			&  92.18		              	& 98.76		               	 \\

\hline
\hline
\multicolumn{5}{c}{Shallow \& Wide Architecture (ResNet-50 2x)}   \\
\hline
\hline

SimCLR 		   						& 94        			& 22			&  91.43               	& 97.71               	 \\
\textbf{SimCLR (SK)}  					& 140       			& 22			&  \textbf{94.44}        & 98.27               	 \\
SwAV		 						& 94        			& 9 			&  93.04               	& 98.73               	 \\

\hline
\hline
\multicolumn{5}{c}{Deep \& Narrow Architecture (ResNet-152 1x)}   \\
\hline
\hline

SimCLR		    						& 58        			& 24			&  92.69               	& 98.02               	 \\
\textbf{SimCLR (SK)}	 				& 89        			& 24 			&  \textbf{93.85}        & 98.43               	 \\ 
Random Weight						& 58  			& 52			&  93.62              	& 98.58               	 \\
\hline
\end{tabular}}
\caption{Snapshot Serengeti: Model fine-tuned with SSL representations.}
\label{table:SSLvsSOTA-Serengeti}
\end{table}

\paragraph{Snapshot Serengeti}
Table \ref{table:SSLvsSOTA-Serengeti} presents our SSL-based results on the Serengeti dataset. We make several observations. 

First, for the three varying architectures, a fair comparison among the models with the same number of parameters (excluding the Selective Kernel based SimCLR as it uses larger number of parameters) reveals that the SSL-based model's top-1 accuracy is competitive to that of the random weight based model. For example, in the shallow-narrow architecture domain (ResNet-50 1x), the DeepCluster-pretrained model achieves top-1 92.95 test accuracy, which is slightly greater than the random-weight based model's 92.18 accuracy. DeepCluster converges to its optimal performance in 10 epochs, which is considerably smaller than random-weight based model's 55 epochs training time. Also in the deep-narrow architecture domain (ResNet-152 1x), the SimCLR-based model converges to the top-1 92.69 test accuracy within 24 epochs, narrowing the gap with the random-weight based model's 93.62 accuracy that it achieved in 52 epochs. 


Second, when we compare the performance of SSL-based top-performing models with the SL-based model's performance, we see that the SSL-based models exhibit slightly better generalizability. For example, the ResNet-50-1x Barlow Twins and DeepCluster models' top-1 test accuracy values are 92.79 and 92.95, respectively, which are slightly greater than the performance of a model with the same capacity that is fine-tuned with the SL representations (i.e., 91.93). Another benefit of the SSL model is that it is more efficient. For example the ResNet-50-1x Barlow Twins and Deep Cluster models converge to the optimal performance approximately 1.6 times faster than the SL model. The Barlow Twins and DeepCluster-based models require 10 and 11 epochs to convergence, respectively, while the SL-based model requires 18 epochs.

Third, increasing model capacity results in improved generalizability. For example, SimCLR benefits from increased width and depth. The top-1 test accuracy values from SimCLR-based ResNet-50-1x (shallow and narrow), ResNet-50-2x (shallow and wide), and ResNet-152-1x (deep ad narrow) models are 90.62, 91.43, and 92.69, respectively. We observe a similar increase in performance for the SwAV-based models, e.g., the ResNet-50-1x (shallow-narrow) and ResNet-50-2x (shallow-wide) models achieve 92.14 and 93.04 accuracy, respectively.

Four, when we use the SimCLR representations optimized with the Selective Kernel technique, it outperforms the random-weight based top-performing model in both the shallow-narrow and deep-narrow architecture domains. Another useful observation is that the wide SimCLR model (ResNet-50 2x) with the Selective Kernel optimization achieves the best generalization performance on the Serengeti dataset, superseding the performance of both the random-weight based model and SL-based model. However, this comes at the cost of a substantial increase in the SimCLR model's parameters (140M parameters of SimCLR vs. 58M parameters of the random-weight based top-performing model). Thus, SSL representations exhibit slightly better generalizability than the SL presentations when the target domain is significantly different from the source domain. \textbf{This observation addresses our RQ1}.

Five, some shallow-narrow (ResNet-50 1x) SSL-based models such as Deep Cluster, Barlow Twins, and SwAV achieve better performance than the SL-based model with the same capacity (all models use 24M parameters) with high efficiency. I.e., the SSL models converge significantly faster, e.g., SwAV requires as low as 6 epochs to converge to its optimal performance. Thus, SSL-based pretrained representations transfer effectively to domains that are very different from the source domain, and adaptation to the novel domain can be done significantly faster. The shallow and narrow architecture domain for the SSL-based models is appealing due to its modest space requirement and expedited fine-tuning time.

Finally, from table \ref{table:SSLvsSOTA-Serengeti} we draw a key insight on the comparative performance of various SSL-based models. We experimented with two SSL approaches: instance-based (SimCLR and Barlow Twins), and clustering-based (DeepCluster and SwAV). We observe that both approaches are comparable in terms of generalizability. The clustering-based approach exhibits better efficiency. For example, SwAV converges within 6 epochs. \textbf{This observation addresses our RQ2}.

\subsection{Study 3}





The goal of study 3 is to determine the level of invariance in the SSL and SL representations for understanding their generalizability. We investigate the invariance property of both models that are fine-tuned using SSL and SL pretrained representations via two sets of experiments. First experiment compares the prediction confidence (i.e., probability for the true class) of both models on artificially transformed images. Second experiment compares prediction confidence of both models on images with naturally occurring variation.



First experiment is done in study 3, and second experiment is done in study 4. For the SSL and SL-based models, we use the fine-tuned ResNet-50 1x architecture for both studies. As the SSL technique, we use Barlow Twins.

\subsubsection{Invariance Analysis: Artificially Created Variation}


We select ten images randomly from the following four categories outlined in \autoref{chap:method}.
\begin{itemize}
    \item Category 1 (zebra): High precision/recall for both SSL and SL.
    \item Category 2 (guineaFowl): High precision/recall for SSL.
    \item Category 3 (ostrich): High precision/recall for SL.
    \item Category 4 (lionMale): Low precision/recall for both SSL and SL.
\end{itemize}


 \begin{figure}[!htb]
\minipage{0.50\textwidth}
  \includegraphics[width=\linewidth]{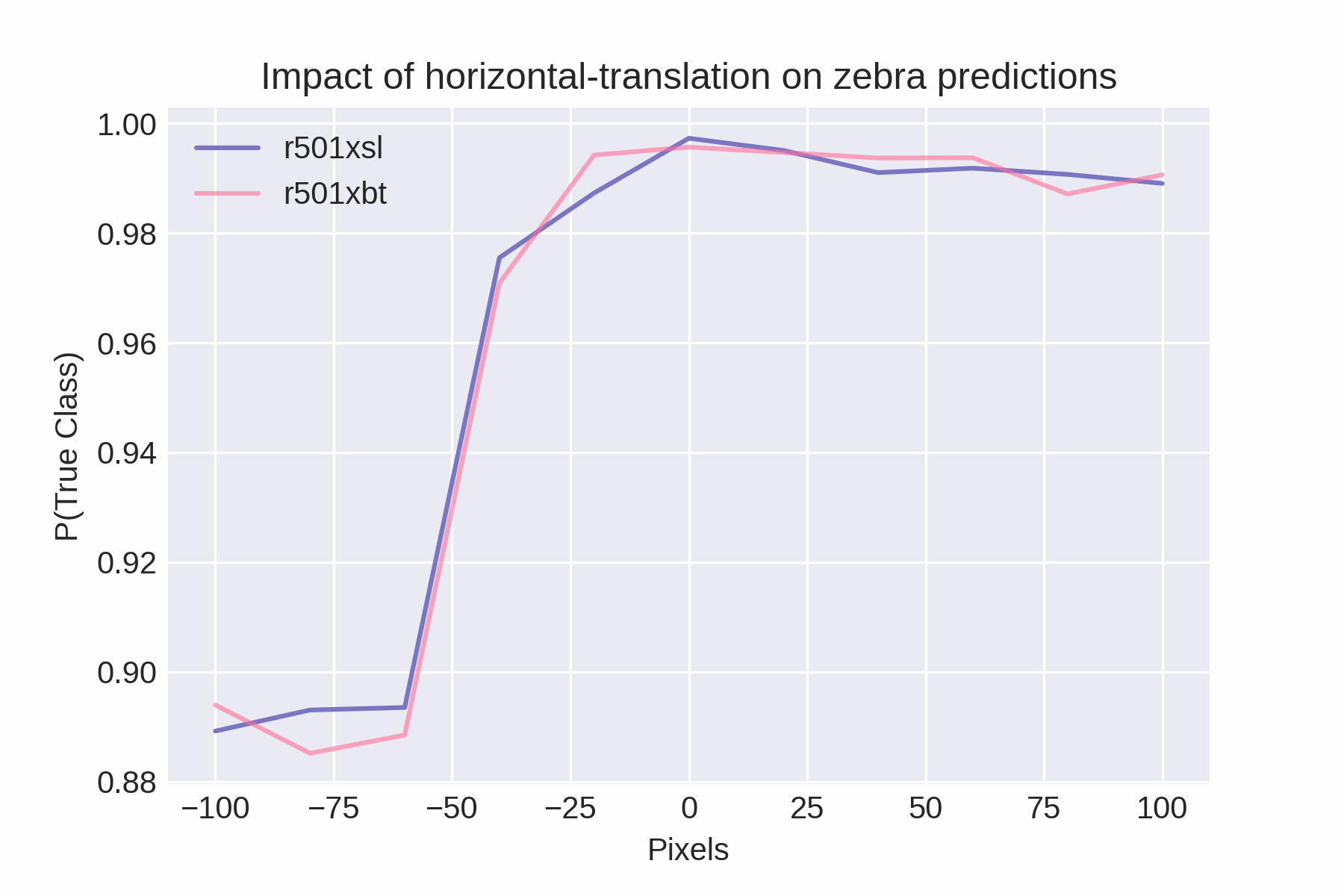}
\endminipage\hspace{1.3pt}
\minipage{0.50\textwidth}

  \includegraphics[width=\linewidth]{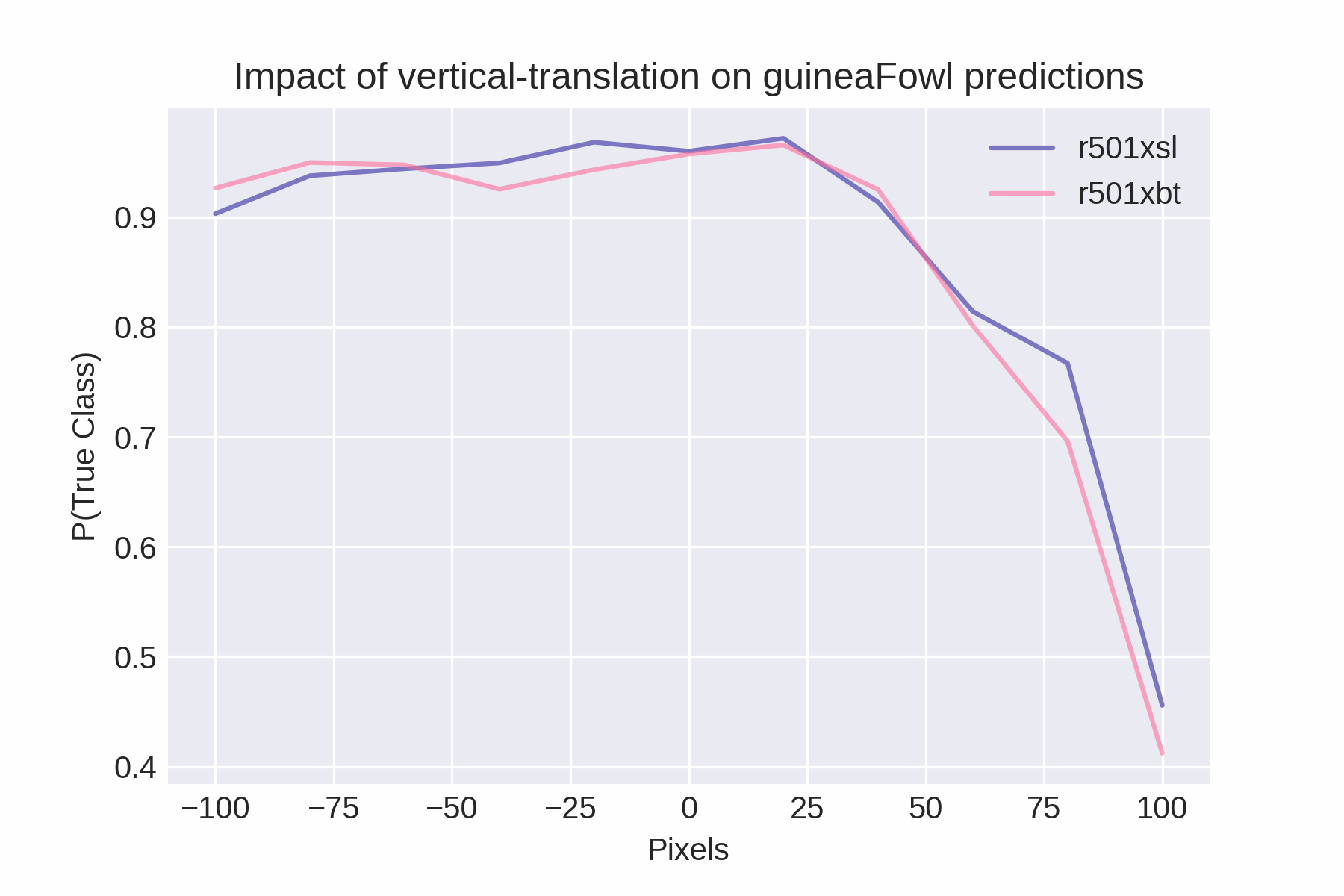}
\endminipage\hspace{1.3pt}

\pagebreak
\minipage{0.50\textwidth}%
  \includegraphics[width=\linewidth]{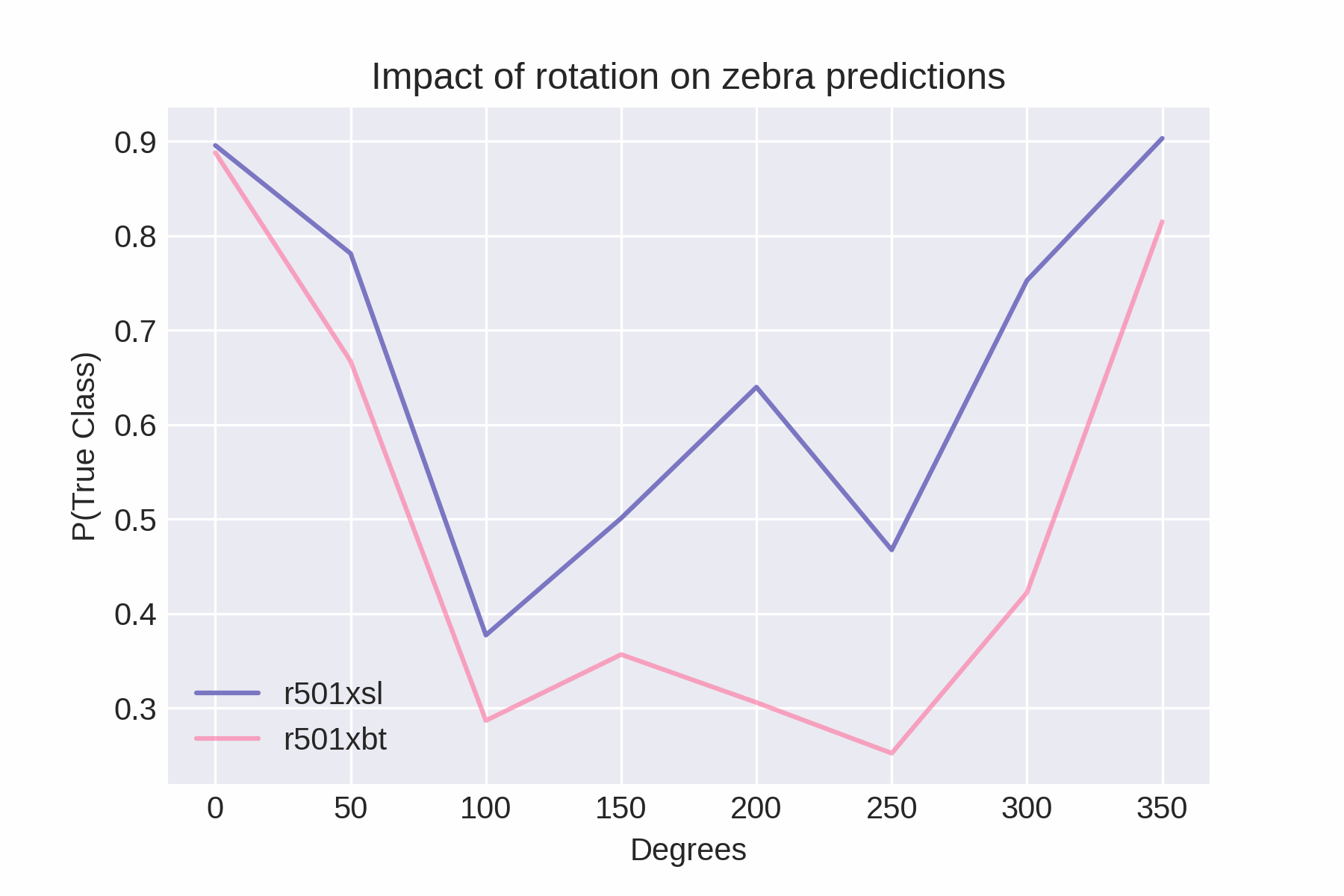}
\endminipage\hspace{1.3pt}
\minipage{0.50\textwidth}%
  \includegraphics[width=\linewidth]{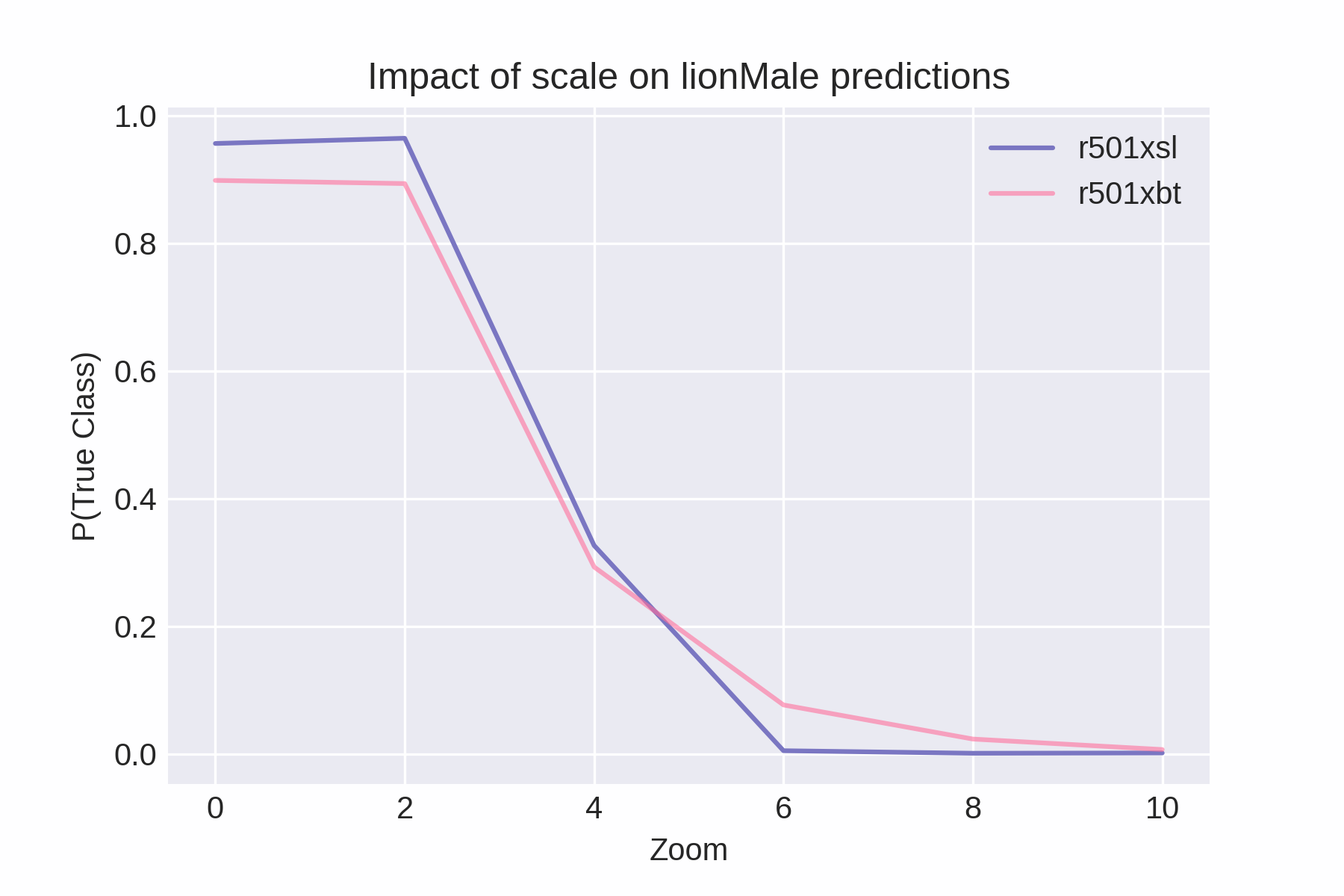}
\endminipage\hspace{1.3pt}
\pagebreak
\minipage{0.50\textwidth}%
  \includegraphics[width=\linewidth]{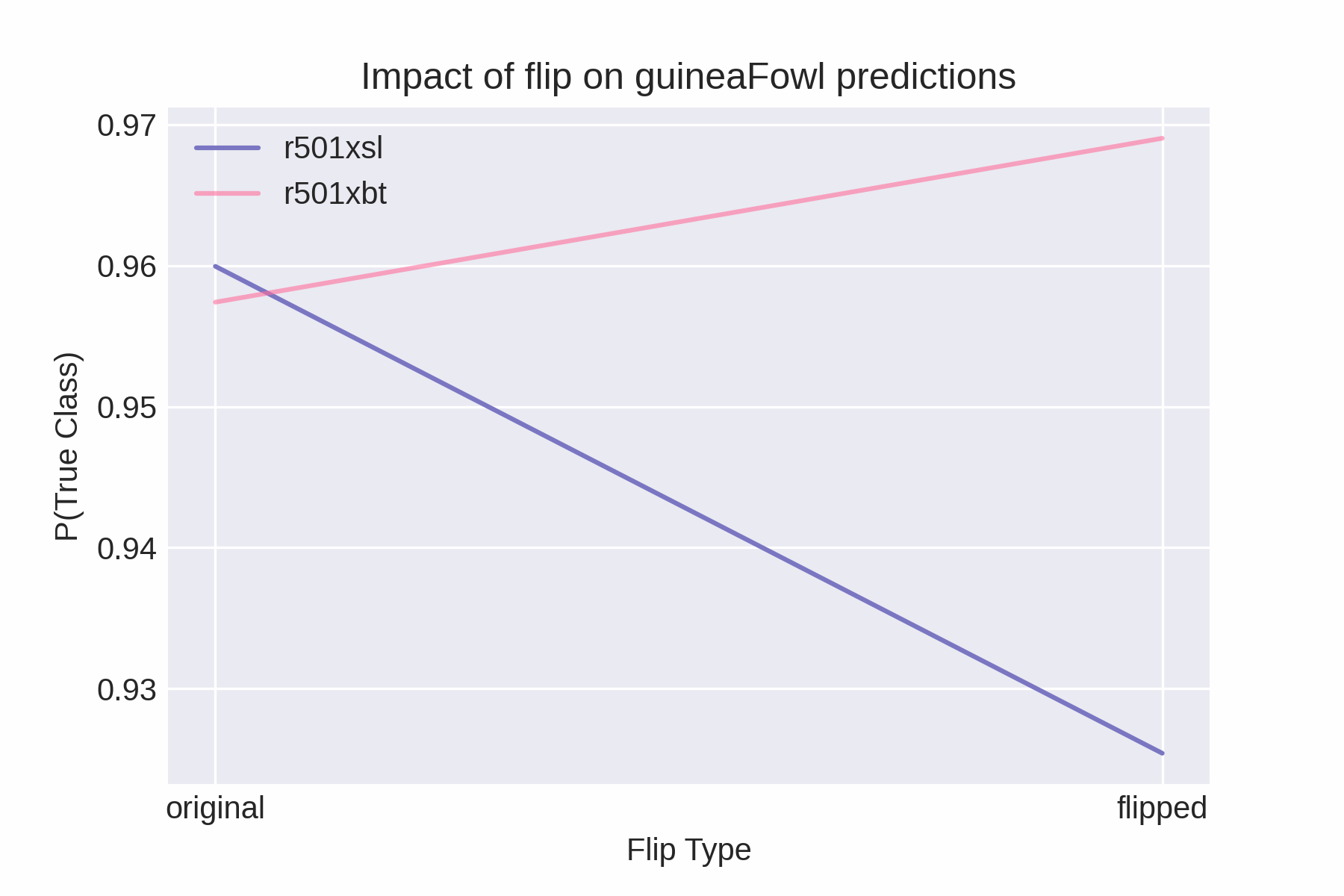}
\endminipage
\caption{Analysis of invariance to various spatial transformations of the images. Blue curve refers to the SL-based model and the purple curve represents the SSL-based model.}
\label{fig:invariance-analysis}
\end{figure}

We perform five transformations on the images from each category: horizontal translation, vertical translation, rotation, scale change, and vertical flip. In all these transformations, the entire input distribution is altered that include the animal and its background. For each transformation, we compute the average prediction probability for ten images. Finally, the average probabilities are plotted against the transformations. 


In general, the SSL and SL-based models perform similarly in the context of invariance to transformations. Figure \ref{fig:invariance-analysis} reports a subset of the plots that are taken from all transformations across each category. Since there is not much variation in the plots, we only show a few for illustration. From these plots we identify a scenario in which the SL-based model performs slightly better (e.g., for rotation transformation on Zebra) than the SSL-based model, as well as a scenario in which the SSL-based model shows slightly better generalizability (e.g., for flip transformation on Guinea Fowl). Results of this experiment suggest that at the aggregate level both the SSL and SL-based models exhibit similar invariance property with respect to some spatial changes in the entire input distribution. \textbf{This observation partially addresses our RQ3}. We still need to understand what happens when the main object (i.e., the animal) in the images go through some spatial changes (e.g., translation, rotation, scale change) as well as when only its background (e.g., field, sky) and global illumination change. We investigate this in study 4 using naturally occurring images.

\subsection{Study 4}

%

The goal of study 4 is two-fold: analyze the invariance of SSL and SL representations using naturally occurring variation in the images as well as the decision making process the SSL and SL-based fine-tuned models. These two goals are accomplished by analyzing the prediction probabilities on a carefully selected set of natural images along with their attributions, i.e., saliency maps of the final convolutional layer of these models. 







%
%



For each image, we compare the prediction probabilities of two models. We seek to understand the confidence of a model about its decision while reasoning about the class of an image against naturally occurring variation in the animals, e.g., translation, rotation, scale variation, as well as in their background and illumination. For capturing the reasoning process, we visualize the part of the input image that is used by a model to determine the class. This is done by generating saliency maps of the final convolutional layer of these model. By using the saliency maps we will be able to learn when a model makes a decision about the category of an object, what type of information (i.e., which part of the input pixel space) it uses. This is helpful for debugging the decision making process of a model. Finally, by comparing the saliency maps and prediction confidence, we analyze a model's invariance property.



This study is done for the following categories of images that capture three types of variations: in the animal's spatial location, in its background and in global illumination. In addition to this, we include a set of images that are not easily discernible, e.g., either the main object (animal/human being) is blended with the background or appears ambiguous. For each category, we select four images from the gold standard test set. Then, we compare the prediction probabilities for the true class made by both models along with their saliency maps. 


\begin{itemize}

\item Images with varying translation

\item Images with various rotation

\item Images with varying scale

\item Images with varying background

\item Images with varying illumination

\item Images that are not easily discernible (main object is blended with the background or appears ambiguous)

\end{itemize}

\begin{figure}[h!]
\begin{centering}
\includegraphics[width=12cm]{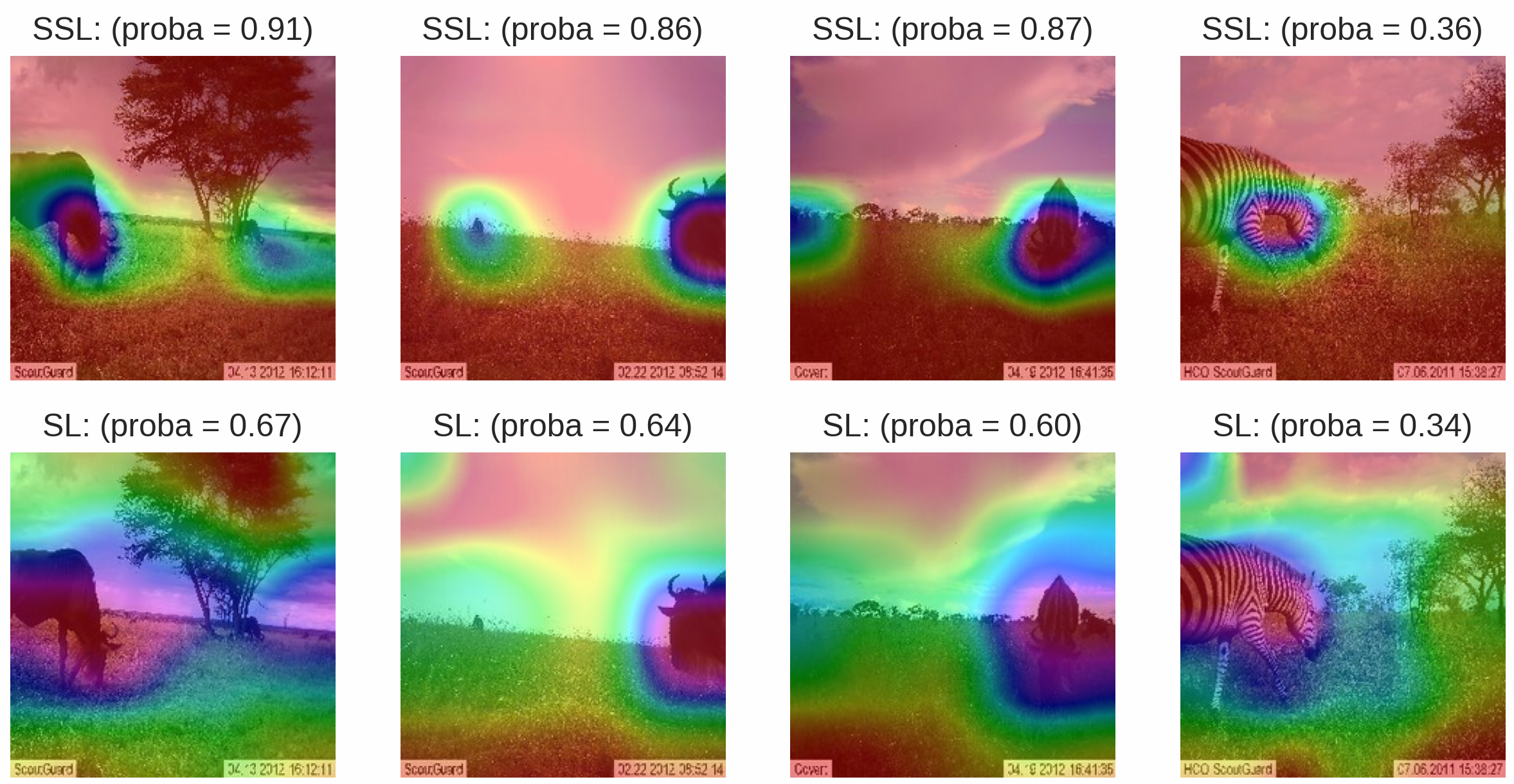}
\caption{Saliency maps of images with varying translation.}
\label{fig:attribution-translation}
\end{centering}
\end{figure}

\paragraph{Images with Varying Translation}

Figure \ref{fig:attribution-translation} shows four images on which both models make accurate predictions about the main object. However, their prediction confidence varies significantly. These are natural images on which animals are not center-positioned. Instead the animals are located either on the far left or right on the frame. We see that the SSL-based model's confidence is considerably larger than SL-based model's confidence on the first three images. We make two observations about the saliency maps. First, the SSL-based model's focus is sharper (i.e., the high-confidence blue and green regions have large intensity). Second, its focus is localized (i.e., the high-confidence blue and green regions are concentrated around smaller areas). This is in contrast to the SL-based model that uses a blurred focus (i.e., green regions are washed away across a large area of the background) and a wide-angle lens (i.e., the high-confidence blue and green regions are concentrated around larger areas). It is the quality of the lens (sharp and small focus) that equips the SSL-based model with the ability to locate the animals in the images more accurately for reasoning about their category. On the last image of Zebra, although the SSL-models predicted probability for the Zebra class is slightly larger than that of the prediction by the SL-based model, the SSL-based model sharply focuses on the stripe pattern on Zebra's neck and upper part of the right front leg. On the other hand, the SL-based model, due to its wide-angle and blurry lens, focuses almost over the entire animal as well as on part of the background. Thus, we see that the reasoning process of both models are significantly different, and that the SSL-based model is more confident about its convictions. Based on the prediction probability and focal region on the input, it seems that the SSL-based model has better invariance with respect to translation in the animals in the images.

\begin{figure}[h!]
\begin{centering}
\includegraphics[width=12cm]{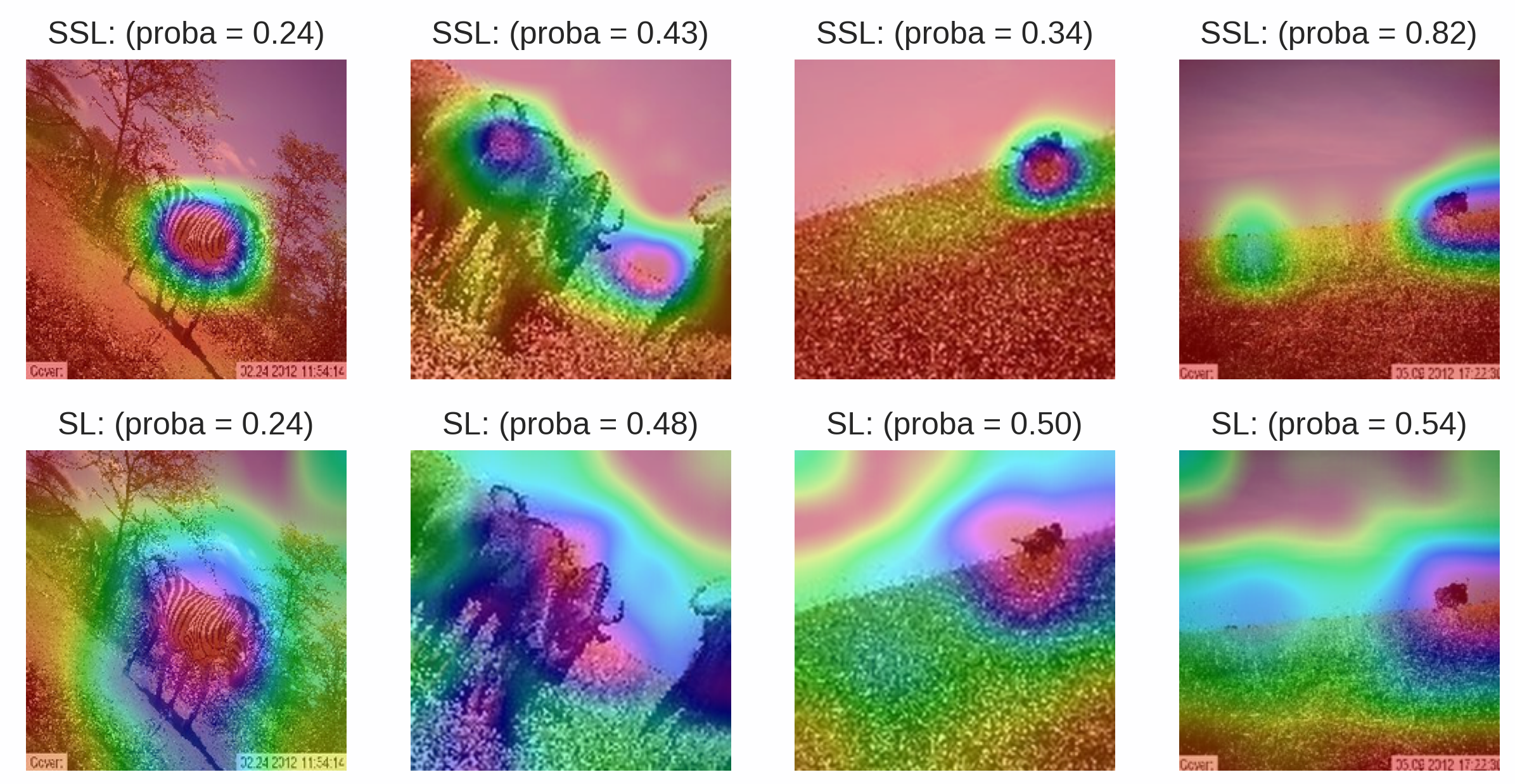}
\caption{Saliency maps of images with varying rotation.}
\label{fig:attribution-rotation}
\end{centering}
\end{figure}

\paragraph{Images with Varying Rotation}

Figure \ref{fig:attribution-rotation} shows four images with varying rotation. Only the first image is incorrectly predicted by both models. These images are unlike the translation-based images (figure \ref{fig:attribution-translation}), because on these images the entire frame is rotated that includes both the animals and their background. Unlike the translation attribution results, the SSL-based model does not seem to exhibit better invariance with respect to rotations of the images. When the rotation angle is larger (2nd and 3rd image), the SL-based model makes predictions with higher confidence. However, consistent with the previous saliency maps, the SSL-based model uses a sharp focus to locate an animal by looking at a smaller region.



\begin{figure}[h!]
\begin{centering}
\includegraphics[width=12cm]{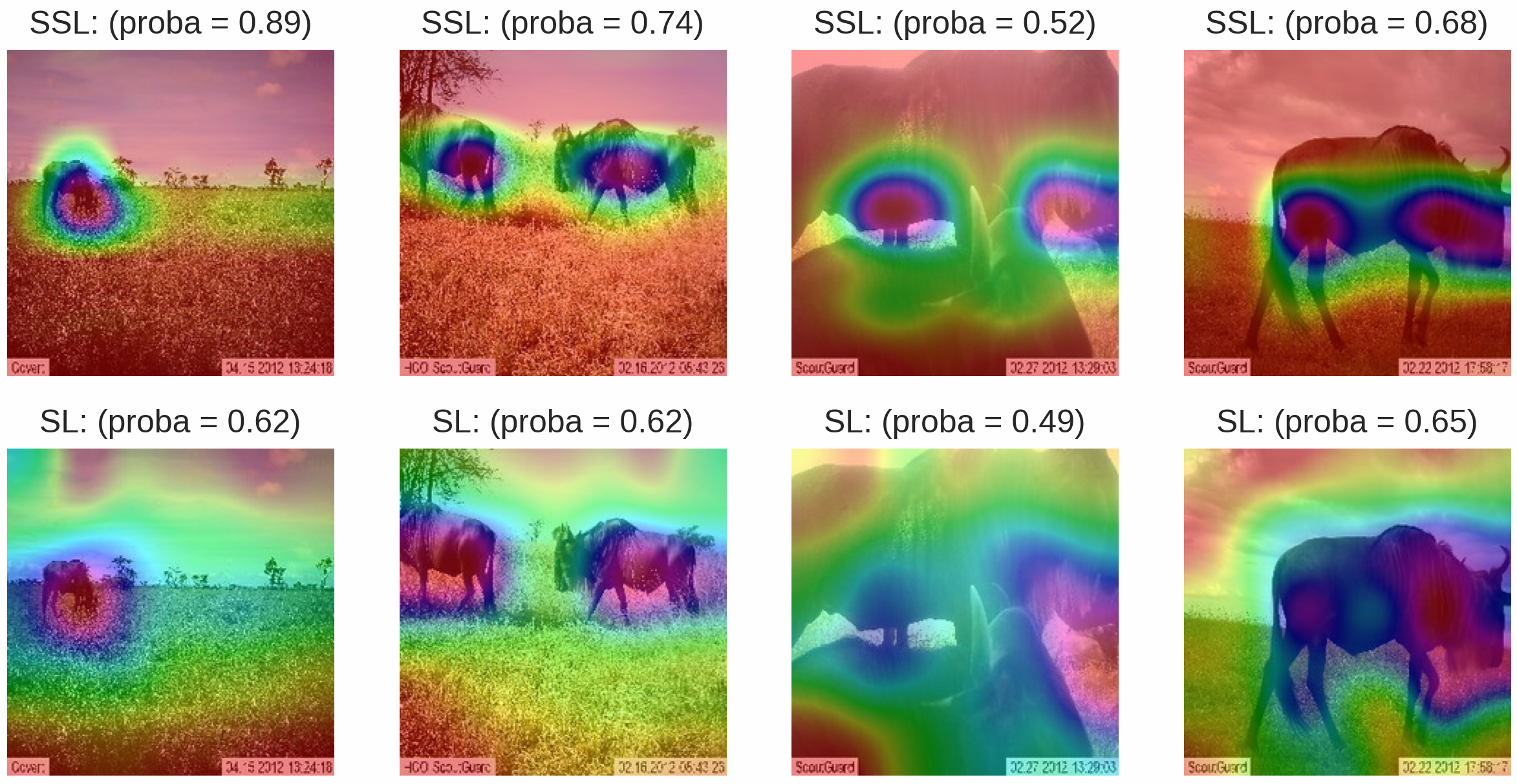}
\caption{Saliency maps of images with varying scale.}
\label{fig:attribution-scale}
\end{centering}
\end{figure}

\paragraph{Images with Varying Scale}

Figure \ref{fig:attribution-scale} shows how the SSL and SL-based models reason about the class of the animals when their scales vary. Both models make accurate predictions about the main object in all four images. Similar to the translation-based attribution figures, SSL-based models have higher confidence on its predictions, which is due to its ability to focus sharply on the distinctive part of the image. We see that when scale of the animal is smaller, the difference between the confidence of the SSL and SL-based model increases, which indicates the SSL-based model's better inavariance with respect to scale variation.

\begin{figure}[h!]
\begin{centering}
\includegraphics[width=12cm]{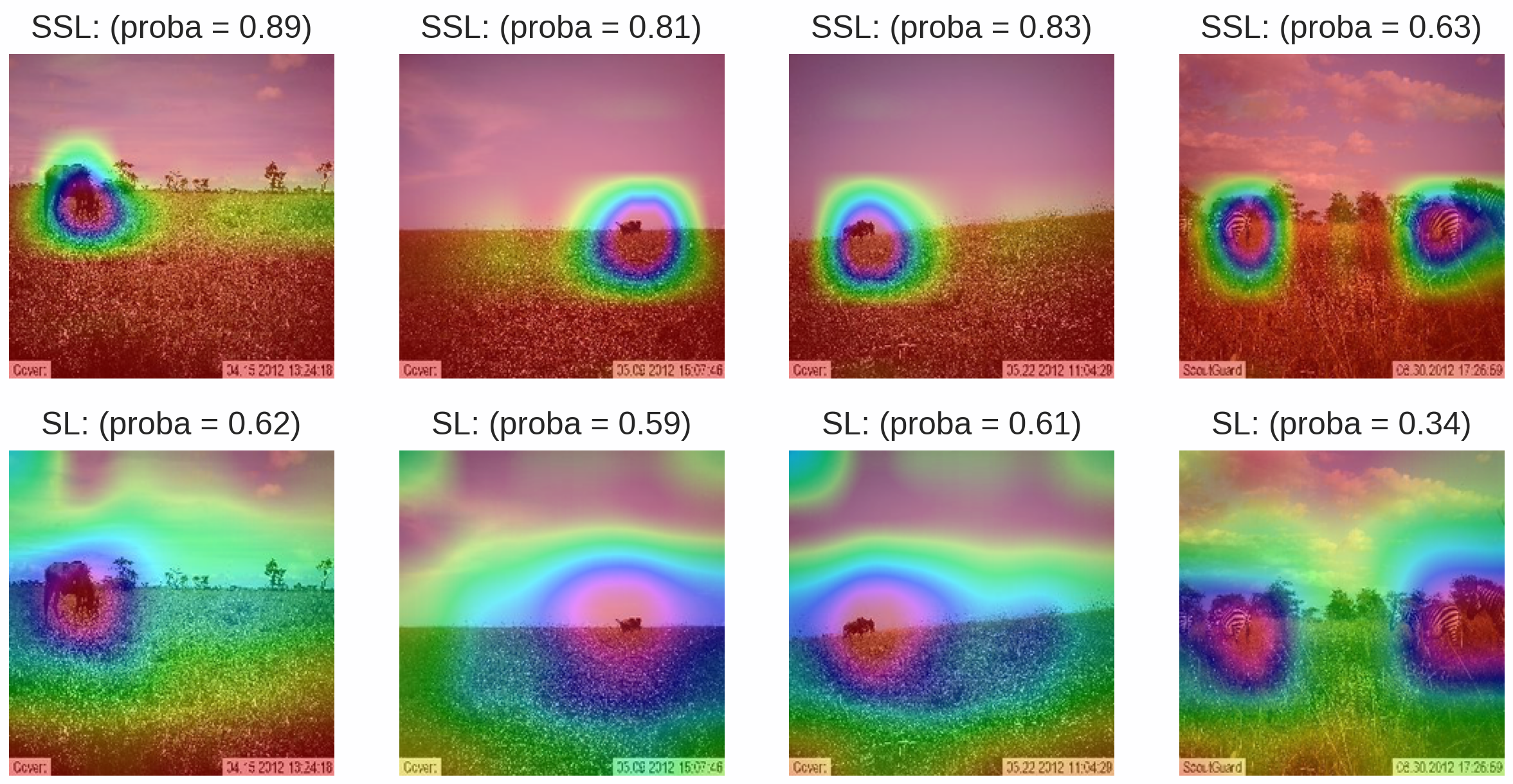}
\caption{Saliency maps of images with prominent background.}
\label{fig:attribution-background}
\end{centering}
\end{figure}

\paragraph{Images with Varying Background}

In the four images in figure \ref{fig:attribution-background}, the background is more prominent and animals are shown in smaller scale. Thus, in these images background occupies most of the frame and the model needs to cope with the variation in the background. Both models make accurate predictions about the main object in all four images. Consistent with the attribution study on scale variation, the SSL-based model's confidence is much larger than the SL-based model. Also, as before, the SSL-based model sharply focuses on the distinctive parts of the animals.

\begin{figure}[h!]
\begin{centering}
\includegraphics[width=12cm]{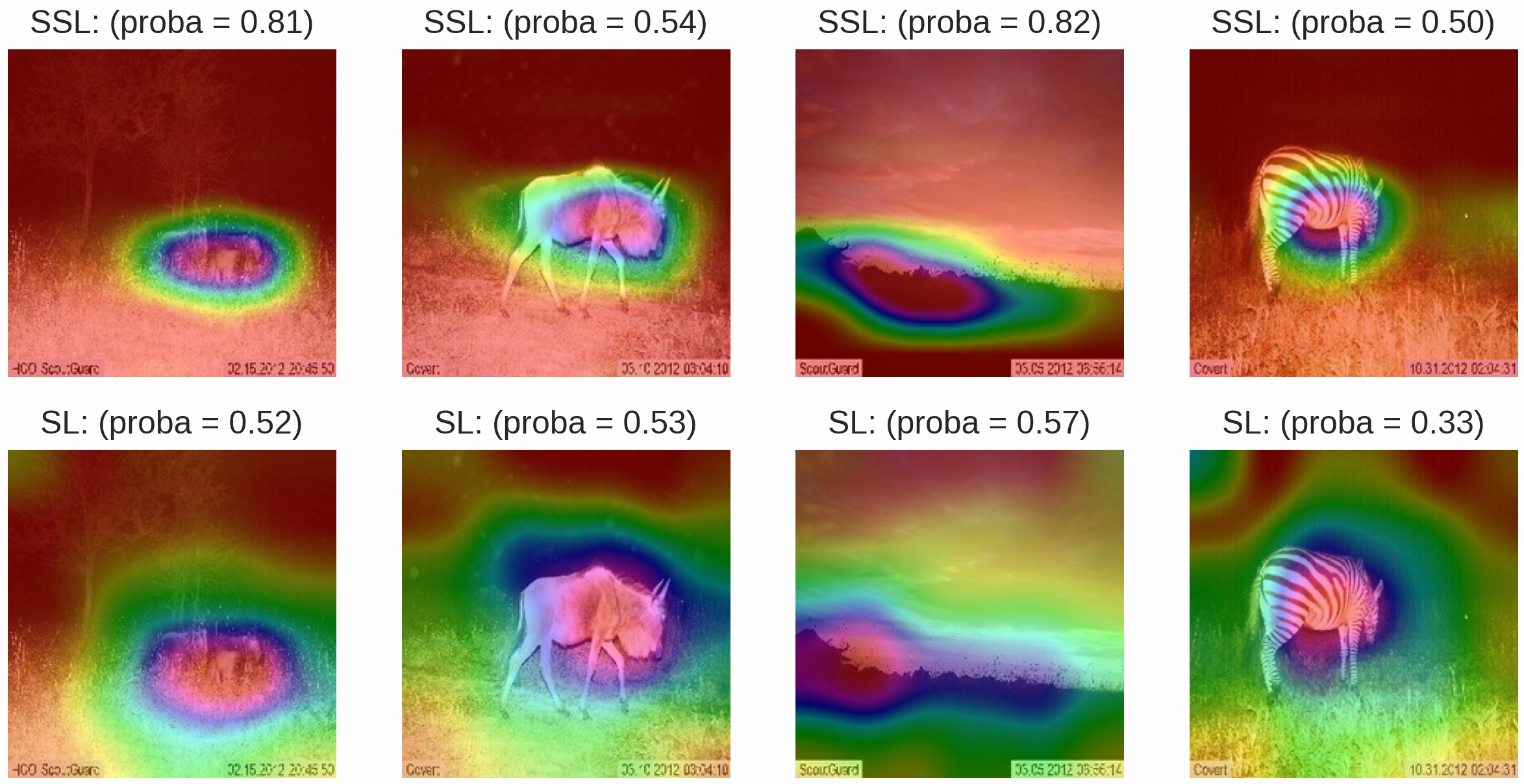}
\caption{Saliency maps of images with varying illumination.}
\label{fig:attribution-illumination}
\end{centering}
\end{figure}

\paragraph{Images with Varying Illumination}

Figure \ref{fig:attribution-illumination} shows four images with varying illumination. Both models make accurate predictions about the main object in all four images. Again, SSL-based model exhibits better confidence in its predictions, which indicates its better invariance against variation in illumination.

\begin{figure}[h!]
\begin{centering}
\includegraphics[width=12cm]{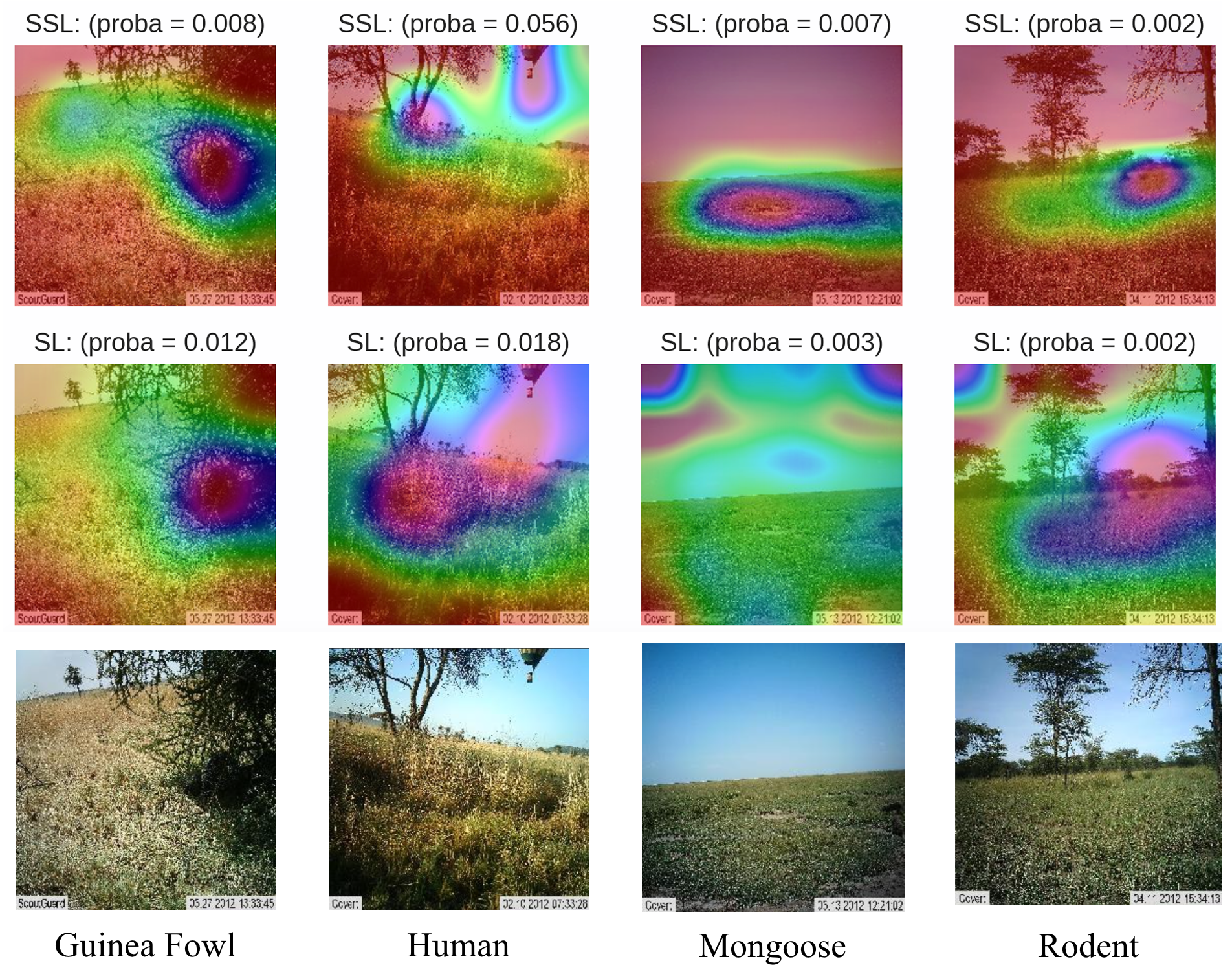}
\caption{Saliency maps of images that are not easily discernible.}
\label{fig:attribution-ambiguous}
\end{centering}
\end{figure}


\paragraph{Images Not Easily Discernible}

Figure \ref{fig:attribution-ambiguous} shows four images that are not easily discernible. Either the scale of the main object is extremely small to the point of being completely indiscernible (human in the 2nd image), or blended with the background (Guinea Fowl in the 1st image, Mongoose in the 3rd image, and Rodent in the 4th image). Both models fail to identify the main objects in these images. However, the SL-based model seems to be able to locate the main object on the first three images. Both models fail to locate the animal on the 4th image, which is understandable from the nature of the image. The animal is invisible to non-expert eyes.

Based on the above analysis, we see that the SSL-based model almost always outperforms the SL-based model in accurately locating the main object on the images. Whenever the SSL-based model exhibits lower confidence in its predictions, it is mainly due to its inability to locate the position on the object. The SL-based model seems to be unable to filter out part of the background while locating the main object. On the other hand, the SSL-based model is very good at ``seeing'' the discriminating part of the input by filtering out unnecessary information that includes both the background and parts of the object. The SSL-based model uses just the right amount of information to reason about the semantic identify of the objects in the images. Its ability to acquire a sharp focus and success in locating discriminating pixels illustrates its better invariance property, and explains its effective generalizability as compared to the SL-based model. \textbf{This observation addresses the RQ3}. The analysis in this study also \textbf{addresses the RQ4} about the decision-making process of the SSL-based model.

\subsection{Discussion}

Previously the generalizability of SSL representations were studied for transfer learning classification task in various domains \citep{chen:2020v1, he:2020, misra:2020, zbontar:2021}. However, there has been no attempt to explain the generalizability using the invariance in latent representations. In addition to this, no investigation was done to understand the variance in prediction confidence of the SSL-based model in comparison to the SL-based model. Based on the results obtained from four studies, we attempt to fill this gap. We identify the domain (e.g., camera-trap) that could benefit more from SSL representations. We explain the efficacy of the SSL-based models by studying their generalizability. We argue that generalizability should not only be evaluated using prediction accuracy, and emphasize the importance of including the model's prediction confidence as well as reasoning process. 

There are diverse techniques for creating self-supervised representations. Previously comparative analysis of the SSL techniques was done \citep{zbontar:2021}. However, the analysis is solely based on the effectiveness (i.e., accuracy). We argue the necessity of including the efficiency (epochs to convergence) for better comparability. In other words, we draw a more holistic picture of the comparative performance of various SSL techniques. We do so by choosing a domain that is significantly different from the source domain, hence challenging for transfer learning.

Some questions still remain though. For example, we used a very large target dataset for understanding the generalizability of the SSL representations. But transfer learning is more beneficial when the labeled target dataset is smaller. Thus, we need to investigate the generalizability of the SSL representations in the low-data regime. Another key issue is whether there exists variability in the invariance of the diverse SSL techniques. For example, how does invariance property vary among the two main SSL approaches, i.e., instance-based and clustering-based?

Network interpretability via attribution is not a novel approach. But no previous studies were done to shed light on the decision making process of the SSL-based models and identify their distinctive characteristics. We show that attribution saliency maps can be used to explain a SSL-based model's invariance property. But our approach was only limited to the attribution of the final convolutional layer. It will be useful to investigate how invariance is encoded in other hidden layers by using feature visualization techniques \citep{yosinski:2015}. Finally, to acquire a deeper understanding of the reasoning process of the SSL-based models, we should look at other types of symmetries that include equivariance and covariance. In this research, we take the first step towards discovering distinctive processes employed by the SSL-based and SL-based models to reason about the semantic identity of the images.

\section{Conclusion and Future Work}


In this article, we conduct a domain-based extensive study to understand the generalizability of the SSL representations in comparison to the SL representations for solving transfer learning classification tasks. We focus on two types of target datasets: similar to the source domain (CIFAR datasets), and significantly different to the source domain (Serengeti camera-trap dataset). We show that SSL representations are more generalizable as compared to the SL representations. To derive this observation, we use prediction accuracy as well as prediction confidence of the SSL and SL-based models. Moreover, by creating saliency maps we analyze the attribution of the final convolutional layer of these models. We explain the generalizability of the SSL representations by studying their invariance property. The saliency maps show that SSL-based models are very good at identifying the most discriminative part of the input to reason about its semantic category. As a result, the SSL-based model's prediction confidence is comparatively larger than that of the SL-based model. By comparing the attribution of the SSL and SL-based models, we show that the SSL-based model exhibits better invariance, which explains its improved generalizability.

As future work, we plan to extend the study on the generalizability of SSL representations. We will create domain-specific representations \citep{Beltagy:2019} to compare them with domain-agnostic representations. We will determine how the generalizability, as well as the invariance property, vary across these two representations. In addition to this, we plan to perform a study on feature visualization to obtain more insights into the reasoning process of the SSL models.


\bibliography{references}

\end{document}